\def\year{2020}\relax
\documentclass[letterpaper]{article} 
\usepackage{aaai19}  
\usepackage{times}  
\usepackage{helvet} 
\usepackage{courier}  
\usepackage[hyphens]{url}  
\usepackage{graphicx} 
\usepackage{todonotes}
\usepackage{graphicx}  
\usepackage[english]{babel}
\usepackage[utf8]{inputenc}
\usepackage{algorithm}
\usepackage{algpseudocode}
\usepackage[backend=biber,style=numeric,sorting=none]{biblatex}
\usepackage{mathtools}
\usepackage{graphicx}
\usepackage{tocbibind}
\usepackage[export]{adjustbox}
\usepackage[toc,page]{appendix}
\usepackage{comment}
\usepackage{color,soul}
\usepackage{siunitx}
\usepackage{booktabs}
\usepackage{makecell}
\newcommand*\rot{\rotatebox{90}}

\pdfoutput=1

\title{Interpretable Machine Learning Models for Predicting and Explaining Vehicle Fuel Consumption Anomalies
}
\author{Alberto Barbado\textsuperscript{\rm 1,2}\thanks{Corresponding author at Telefónica, 28050 Madrid, Spain. E-mail address: alberto.barbadogonzalez@telefonica.com (A. Barbado)} , Oscar Corcho\textsuperscript{\rm 2}\\
alberto.barbadogonzalez@telefonica.com, ocorcho@fi.upm.es\\
\textsuperscript{\rm 1} Telefónica, 28050 Madrid, Spain\\ 
\textsuperscript{\rm 2} Departamento de Inteligencia Artificial, Universidad Politécnica de Madrid, 28660 Boadilla del Monte, Spain\\ 
}
 
\begin{document}

\maketitle

\begin{abstract}
Identifying anomalies in the fuel consumption of the vehicles of a fleet is a crucial aspect for optimizing consumption and reduce costs. However, this information alone is insufficient, since fleet operators need to know the causes behind anomalous fuel consumption.

We combine unsupervised anomaly detection techniques, domain knowledge and interpretable Machine Learning models for explaining potential causes of abnormal fuel consumption in terms of feature relevance. The explanations are used for generating recommendations about fuel optimization, that are adjusted according to two different user profiles: fleet managers and fleet operators.

Results are evaluated over real-world data from telematics devices connected to diesel and petrol vehicles from different types of industrial fleets. We measure the proposal regarding model performance, and using Explainable AI metrics that compare the explanations in terms of representativeness, fidelity, stability, contrastiveness and consistency with apriori beliefs. The potential fuel reductions that can be achieved is round 35\%.

\end{abstract}
\textbf{Keywords} Explainable Artificial Intelligence. Fuel Consumption. Anomaly Detection. Explainable Boosting Machine. Feature Relevance. Generalized Additive Models.

\section{1. Introduction}
Combining Advanced Analytics techniques together with IoT (Internet of Things) data offers many possibilities to find and extract relevant insights for business decisions. At Telefónica, for instance, we see how the union of Machine Learning (ML) with IoT data helps to create new use cases for the Fleet Management Industry. 
An example of it is the usage of ML for anomaly detection of the fuel consumption of vehicles. For a fleet manager, it is very useful to be able to find which vehicles are having an abnormal fuel consumption, since it is crucial for optimizing costs.

However, detecting which vehicles have an anomalous fuel consumption alone is not enough. Only providing that information leads to more questions than answers. Why are the vehicles consuming that extra amount of fuel? How could it be reduced?. These questions are not answered by a binary output that indicates which consumption are anomalous and which ones are not. 

This is one of the reasons why Explainable AI (XAI) is relevant: it enhances that initial information with different types of explanations, helping to answer those questions that may arise. In fact, XAI is one of the core elements of Responsible Artificial Intelligence (RAI) \cite{alej2019explainable}.

Nonetheless, XAI is still an emerging field with many uncharted or relatively new territories. For instance, how do we now that the explanations generated are good enough? How do we compare different XAI techniques quantitatively in order to find the one that provides better explanations? These questions address the importance of metrics for XAI for measuring the understandability of explanations.

Together with those questions, another issue is the following one: Do the explanations adapt to the user profile? Are they adjusted in such a way that the target audience finds them clear and useful enough?

Also, even though explanations themselves are useful, there is always a caveat present: What happens when explanations contradict apriori knowledge of a field? How do we ensure that apriori knowledge and explanations are aligned?. Though, regarding the first question, it may be possible that explanations differ from domain knowledge either because it is wrong or because it may complement it, in many cases the important question is the second one: ensuring alignment between apriori knowledge and explanations.

Finally, even considering good understandable explanations that are aligned with domain knowledge and that are expressed in an understandable way for their audience, there are still questions unanswered. For example, what shall we do about it? The prescriptive dimension also arises, remarking the importance of not only providing insights, but also suggesting possible actions to further help the decision maker.

Taking all these questions in consideration, in these paper we will propose a complete process to address the business need of not only detecting anomalies within the fuel consumption of a fleet of vehicles, but also explaining what causes them. This process will include how to adjust the explanations to be understandable by its audience, how to include business rules in order to ensure that they are aligned with domain knowledge, and how to provide recommendations on what may be done to reduce the fuel consumption of outliers in order to turn them into inliers. 

We will analyse how to generate these explanations for unsupervised anomaly detection using surrogate models. These models help to find the feature relevance relationship between input features and a target one within the context of the output of the unsupervised anomaly detection.
These surrogate models will include different types of Generalized Additive Models (GAM), which are efficient interpretable algorithms that are able to both model complex non-linear relationships while providing explanations about them. In particular, we will use Explainable Boosting Machine (EBM) \cite{nori2019interpretml}, and analyse its limitations. Within this context, we will also propose a variation over EBM that solves on of those limitations, and we will also analyse a novel GAM algorithm, Constrained Generalized Additive 2 Model with Consideration of Higher-Order Interactions (CGA2M+) \cite{watanabe2021cga2mplus}, which is able to solve other of the limitations with EBM.

We will benchmark these models from a complete point of view that considered both metrics for model performance, as well as metrics to quantitatively analyse the XAI dimension. This approach follows the principles of Responsible AI by Design that considers and includes XAI from the beginning of a ML model life cycle \cite{benjamins2019responsible}.

Following this, the main contributions of our work are:
\begin{itemize}
    \item Proposing a complete methodology for unsupervised anomaly detection in the fuel consumption of the vehicles of a fleet, which includes: data gathering and processing from input sources (telematic devices), detecting outliers in an unsupervised manner, generating explanations of what affects the fuel consumption of outliers, aligning those explanations to business rules, generating recommendations of what may be done to turn outliers to inliers, and adjusting them to be easily understandable for their target audience, considering two different user profiles that may benefit from them. This proposal is based on our patent \cite{patent2020fleet}.
    \item Proposing a variation over EBM algorithm ("EBM\_var") that takes into account a set of categorical features in order to adjust the predictions and features importance.
    \item Comparing the results from EBM, CGA2M+ and EBM\_var both regarding model performance, as well as using XAI specific metrics in terms of representativeness, fidelity, stability, contrastiveness and consistency with apriori beliefs. Regarding CGA2M+, our proposal is, to the best of our knowledge, the first evaluation of that algorithm with real-industry data.
\end{itemize}

The rest of the paper is organized as follows. First, we describe some related work in the area of anomaly detection for fuel consumption, together with previous works regarding XAI for feature relevance explanations. This Section describes other works regarding the combination of explanations with domain knowledge, as well as some of the research conducted regarding metrics for measuring explainability. Then, we describe the different steps of our process proposal, including the metrics for comparing the different model alternatives, also measuring the quality of the explanations. We will also include indicate what business rules are we considering for expressing the domain knowledge and how to combine them with the explanations. Following  this,  we present an empirical evaluation and comparison using real-world IoT data. We then conclude showing also potential future research lines of work.

\section{2. Related Work}

\subsection{2.1. Anomaly detection for fuel consumption}
The detection of anomalous fuel consumption in vehicles from a fleet is present at different research works within the literature. In \cite{aquize2017self}, the authors show how to detect fuel anomalies using unsupervised algorithms (Self-Organizing Maps, SOM). The authors aim to find fuel fraud situations within fleet vehicle data at Bolivia (using a data set of 1000 vehicles with 190627 data points). These situations are normally linked to high fuel purchases within a short period of time. They effectively show how to find clusters within the space of the SOM in order to identify fuel anomalies and detect fraudulent scenarios by evaluating their proposal over a test set. As the authors mention, there are many features that can be used in order to contextualize the fuel consumption (p.e. the normal monthly consumption of the vehicle, the behaviour of other vehicles of the same subgroup...). Their proposal leads only to an output that identifies anomalies, but it could be greatly enhanced with XAI techniques that provide additional insights on what contextual features are relevant for that high fuel consumption.

Fuel fraud is not the only case of possible fuel anomalies within a fleet. As described in \cite{zhang2017safedrive}, driving behaviour may also lead to an increased fuel consumption. Within driving behaviour variables they mention several features, such as RPM speed, acceleration (both forward, and negative from braking), over speed or gear position.

Even though the previous literature includes researches related to the detection of anomalous fuel consumption (both from fraud scenarios and from contextual variables), to the best of our knowledge there are no previous works regarding the explanation of those anomalies using XAI techniques.

\subsection{2.2. Surrogate methods for feature relevance}
From among the different outputs that a post-hoc XAI technique can provide, one of them is in terms of feature relevance \cite{molnar2019interpretable, alej2019explainable}. This output quantifies the individual contribution of each training feature to the target variable. This type of explanations are the ones that we need for our proposal, since we want to explain in term of individual feature contributions how each variable affects the target one for a specific vehicle and date. Thus, along with feature relevance, we are interested in local explanations. They can be provided both by posthoc techniques over the ML model, as well as directly from some whitebox ones. For our analysis, we will focus on interpretable models since they are able to offer both individual and global explanations. The advantage of these models is that they do not need an additional XAI technique over them to infer the explanations; they are directly provided by them.

\subsection{2.2.1. Generalized Additive Models}
In \cite{alej2019explainable} authors proposes a guideline for ensuring interpretability in AI models, indicating that a whitebox algorithmic model should be tried before considering blackbox+XAI combination. The literature is advancing on the research of whitebox models that have performances on pair with complex blackbox ones, in order to contribute to the usage of models that do not need posthoc XAI techniques to understand how they took a decision. This is the case of Generalized Additive Models (GAM) \cite{hastie1987generalized}. In GAM models, the input features and the output have an additive relationship, with each term contributing independently. Therefore, they can be used for knowing the individual impact of each feature in the output for a particular feature value. This idea is similar to Linear Regression models, but the main difference is that the individual relationship between a feature and the output is not constant; is a function that may even be non linear. (Equation \ref{eq:GAM}) shows the basic structure of the GAM algorithm, with $\beta_0$ being the intercept, $i$ a particular feature, $x_i$ its corresponding feature value, and $f_i$ the function that models the relationship with the output. As we see, the function is additive, summing the contribution of the different features in order to obtain the expected output. $g(E[y])$ represents the result of the link function applied over the output feature $y$.

\begin{equation}\label{eq:GAM}
g(E[y]) = \beta_0 + \sum_{n=1} f_{i}(x_{i})
\end{equation}

The previous GAM is improved by $GA^2M$ algorithm \cite{lou2013accurate}, which is the algorithm behind Explainable Bososting Machine (EBM). The difference between them is that EBM is a faster implementation of it. EBM have several improvements over the original GAM. First, the feature functions of EBM can be learned through bagging and boosting techniques. During boosting, only one feature is trained at each step (round-robin) using a very low learning rate in order to make the feature order used irrelevant. This round-robin procedure also lessens the effects of colinearity. Finally, if there are pairwise interactions between features, EBM can detect them and include them as additional terms, as shown at (Equation \ref{eq:EBM-pairwise}) \cite{nori2019interpretml}. There we see that along with the part corresponding to the previous GAM equation, the algorithm also includes the pairwise terms through $\sum f_{ij}(x_{i}, x_{j})$, which models the joint contribution of feature $i$ with feature $j$ through an additional function $f_{ij}$.

\begin{equation}\label{eq:EBM-pairwise}
g(E[y]) = \beta_0 + \sum f_{i}(x_{i}) + \sum f_{ij}(x_{i}, x_{j})
\end{equation}

An additional evolution over the previous algorithm is Constrained Generalized Additive 2 Model with Consideration of Higher-Order Interactions (CGA2M+) \cite{watanabe2021cga2mplus}. CGA2M+ includes two improvements over EBM. First, it allows to specify monotonic constraints, so the functions that model the relationship between an input feature and the output may be monotonic. Second, they model allows to use higher-order interactions, as opposed to EBM, where the interactions are limited to second-order.The hyperparameters of the algorithm allows specifying the features that should include monotonic constraints in order to consider them during the training of the model. Then, the algorithm models the indidivual relationship between an input feature and the output using a Light Gradient Boosting (LightGBM) algorithm, which may be constrained to be monotonic for that feature if it was specified in the input parameters. (Equation \ref{eq:GA2M+}) shows the CGA2M+ algorithm, allowing to include higher-order terms that model the relationship between more than two features $x_i, ..., x_k$.

\begin{equation}\label{eq:GA2M+}
g(E[y]) = \beta_0 + \sum f_{i}(x_{i}) + \sum f_{ij}(x_{i}, x_{j}) + f_{high}(x_{i},..., x_{k})
\end{equation}

\subsection{2.3. Domain knowledge combined with XAI}
Within the review of \cite{alej2019explainable}, one of the open research challenges is combining domain knowledge with the explanations generated in order to enhance the user's understandability. The review mentions that this challenge is specially addressed through the combination of deep learning blackbox models (connectionism) together with symbolic approaches, that are algorithmic transparent and generally directly interpretable, and with domain knowledge expressed through ontologies. This is the case of \cite{confalonieritrepan}, where the authors propose a variant of the TREPAN algorithm that uses domain ontologies in the XAI phase. TREPAN uses surrogate decision trees to explain any blackbox model (model agnostic). However, as the authors highlight, often those trees are not understandable by a final user. That is why they propose a variation on the algorithm that gathers information from a domain ontology, and uses it to prioritise using features for the splits that are more general within the ontology. The prioritisation is done by penalizing more the Information Gain value from considering a feature for the split if that feature is too specific. They assessed their proposal with expert users in either finance or medical domain, each of them receiving explanations based on a model trained in a dataset from their area of expertise. They found that indeed using domain knowledge enhances user understandability.

Domain knowledge can indeed be applied to adjust the explanations generated, and it can be done at different moments during a ML model life cycle. It can be done at the ML model itself (for instance, finding hyperparameters that enhance the model understandability), or during the training of a posthoc XAI method. Finally, it can be also applied after the XAI method generates the explanations, in order to adjust them to the existing domain knowledge.

\subsection{2.4. Metrics for XAI}
The review of \cite{alej2019explainable} identifies necessity of metrics to assess the understandability of the explanations generated. The authors propose the following definition of explainability: "Explainability is defined as the ability a model has to make its functioning clearer to an audience". However, there is a generalized lack of metrics to measure how well the explanations generated by different approaches match that definition of explainability. Some aspects that should be considered within those future metrics are "goodness", "usefulness" and "satisfaction" of the explanations, along with the improvement of the mental model of the user thanks to them. Also, it should be measured how the explanations impact the "trust" and "reliance" by the user in the model.

A relevant research on XAI metrics is \cite{carvalho2019machine}, where authors propose a taxonomy of properties for different explainability scenarios that depend on the use case and the audience of the explanations. The scenarios are explanations at a general level, individual explanations, and explanations that are human-friendly. Metric properties regarding general and individual explanations aim to measure the explanation's understandability regardless of the user. Human-friendly ones take the user into account for assessing the explanations. Thus, the properties of a metric to evaluate individual explanations should include aspects like "representativeness" (instances covered by the explanations), "fidelity" (how well the explanations approximate the underlying model), or "stability" (how similar are explanations for similar instances). In contrast, metric properties to check if explanations are human-friendly include aspects like "contrastiveness" (if explanations are in the form of "if you have done X instead or Y, the output would have changed from A to B") or "selectivity" (if explanations do not include all the causes, but only the most relevant ones). 

Another taxonomy for explanation metrics is described in \cite{hoffman2018metrics}. The authors also propose a split between metrics with and without considering the user. First, they refer to "explanation goodness" for metrics that assess explanation understandability regarding the ML model. Within these types of metrics they include properties like "precision".  However, even when a set of explanations have good metric values for "explanation goodness", they may not help users. This is why there is a second group of metric properties for "explanation satisfaction", that include aspects such as "understandability", "completeness", "usefulness" or "feeling of satisfaction". The authors propose a set of questionaries to evaluate all these aspects within explanations.

Though it is true that the use of questionaries is an approach to evaluate the aforementioned metric properties, one of the challenges is turning them into quantitative metrics for automatically assessing the explanations generated by XAI over a ML model. This is something that the literature is already addressing. The work of \cite{melis2018towards} shows how to use quantitative metrics for measuring some of the properties mentioned before. They first consider three families of metrics, "explicitness", "faithfulness" and "stability". Then, they propose different algorithms to infer them, evaluating the results over different data sets. (Table \ref{table:taxonomy_sota}) contains a summary of all these properties.

The research at \cite{barbado2019rule} also shows how to implement different metrics for quantitative measurement of the understandability of explanations. They use four families of metrics, "comprehensibility", "representativeness", "stability" and "diversity". These metrics are calculated over local explanations used for explaining a blackbox model for anomaly detection, where the explanations considered are only focused on the outliers (in order to explain how to turn them into inliers). However, some of the metrics are "explanation specific", since they cannot be used for every type of explanation. The authors generate explanations using rule extraction techniques, and metrics like "diversity" measure the degree of overlapping between the hypercubes generated. Hence, they only work for a particular type of explanations: local explanations through rules. Other metrics, such as the ones for "stability", that measure how many similar data points are classified within the same category, are "explanation agnostic" since they can be easily applied for other type of explanations, such as the case of feature relevance.

Finally, \cite{confalonieritrepan} also include explanations for measuring the model understandability for decision trees, in terms of number of interior nodes and number of leaves, and for measuring user understandability, through using online surveys and registering different metrics. These last metrics include the response time for a user to understand the decision tree, as well as if the users are able to predict the decision tree prediction for an individual data point, and their perceived understanding of the model through a rating given by them. 

\begin{table}[]
\begin{tabular}{lllll}
\hline
                                               &  {\cite{carvalho2019machine}} & {\cite{hoffman2018metrics}} & {\cite{melis2018towards}} \\ \hline
Accuracy                                                    & X        & X        &          \\
Fidelity/Faithfulness                                          & X        &          & X        \\
Consistency                                                   & X        &          &          \\
Stability                                                      & X        &          & X        \\
Comprehensibility                             & X        & X        &          \\
Certainty                                                       & X        &          &          \\
Importance                                                      & X        &          &          \\
Novelty                                                         & X        &          &          \\
Representativeness                                              & X        &          &          \\
Contrastiveness                                                 & X        &          &          \\
Selectivity                                                    & X        &          &          \\
Social                                                          & X        &          &          \\
Focus on the abnormal                                           & X        &          &          \\
Truthful                                                       & X        & X        &          \\
Consistent with apriori beliefs                  & X        &          &          \\
General and probable                                            & X        &          &          \\
Explicitness                                                    &          &          & X        \\
Feeling of Satisfaction                                        &          & X        &          \\
Usefulness                                                     &          & X        &          \\
Completeness                                                    &          & X        &          \\
Sufficiency of Detail                                           &          & X        &          \\ \hline
\end{tabular}

\caption{Summary of the metric properties within the literature referenced, including some direct mappings between them.}
\label{table:taxonomy_sota}
\end{table}

\section{3. Method}
In this Section we will describe our proposal for the dynamic generation of explanations applied to anomaly detection of fuel consumption. We will first describe the overall process, and then we will focus on each of the main steps. 

\subsection{3.1. Process overview}
The overall process is described in (Figure \ref{fig:flowchart-fuel-recsys}) in (Subsection A.3 of the Appendices). It contains two main phases, the training phase and the explaining phase. Before going though any of those phases, the process first combines newly arrived data with previous historical (if exists), and then applies a preprocessing step (Subsection 3.2). This step generates a base data frame that we will refer to as FAR (Fleet Analytics Record). Its structure detailed in (Subsection A.2 in the Appendices). It is both used for training the ML model and for detecting the vehicle-dates combinations (data points) that have an anomalous fuel consumption in that day. 
The FAR is then used in the next step for identifying those data points that are anomalous, and provide a visual explanation using a value limit to distinguish inliers from outliers (Subsection 3.3).

After that, the process either applies the training phase for creating a new ML model, or applies the explaining phase, using one previously trained. 
For the training phase, the process trains an interpretable ML model (Subsection 3.4) and obtains its metrics in terms of model performance (Subsection 3.8.1). 

For the explaining phase, the process loads the ML model already trained, and uses it for generating explanations over the new data. They are combined with business rules in order to assure a minimum explanation quality (Subsection 3.5). The explanations are stored and combined with previously generated ones. Then, they are used for generating daily recommendations that show the potential fuel that could be saved for each vehicle (Subsection 3.6).

There are two types of audiences considered for the explanations, one fleet operators that receive the information about individual vehicles that have anomalous fuel consumption, together with recommendations that can be applied for reducing it. Second, fleet managers that receive general explanations about the impact of driving behaviour features in the fuel consumption of the whole fleet, as well as information about the fuel consumption of vehicle models without taking into consideration the extra amount caused by the inefficient driving style (Subsection 3.7).
Figure \ref{fig:ExplanationsDashboardExample} shows the final output explanations for those two user profiles.
The explaining phase also includes XAI metrics that can be used both for comparing the explanations generated by different models, as well as for measuring their quality by themselves (Subsection 3.8.2).

\subsection{3.2. Data preprocessing}
\textbf{Obtain daily features}\\
The first step within the preprocessing module is obtaining the daily aggregated information for each of the vehicles within the fleet. The IoT devices within the vehicles provide real-time information of the vehicle's status. A sample of these raw data with a csv structure can be seen in (Table \ref{table:sample_structure}), and is also available at \cite{xaifuelabg2020}.

\begin{table}[h!]
\resizebox{\columnwidth}{!}{%
\begin{tabular}{llll}
\textbf{time\_tx} & \textbf{vehicle\_id} & \textbf{variable\_id} & \textbf{variable\_value} \\ \hline
2020-10-31 00:02:34.073000+00:00 & b123 & EngineSpeed  & 1200 \\
2020-10-31 00:12:34.073000+00:00 & b124 & VehicleSpeed & 55   \\
2020-10-31 01:12:34.073000+00:00 & b125 & EngineSpeed  & 1200 \\
2020-10-31 02:02:34.073000+00:00 & b124 & TripFuel     & 3.1 
\end{tabular}
}
\caption{Sample of the received data from the IoT devices}
\label{table:sample_structure}
\end{table}

However, for our proposal, we are interested in a daily vision of the vehicle. This is mainly done because the recommendations for the user profiles should have a daily granularity level, and also because the databases for training the models are big enough with this granularity for training a ML model. 
Thus, we aggregate the raw information into a set of features, described at (Subsection A.2 in the Appendices). The features chosen correspond to a business a priori knowledge, since they may affect a vehicle's fuel consumption \cite{zacharof2016review}. These features appear within the literature as potential causes of increased fuel usage both from the driving behaviour influence in fuel economy \cite{zhang2017safedrive}, as well as from the vehicle status and exterior conditions \cite{zhou2016review}. The features have been proven useful for predicting fuel consumption with ML models \cite{9072728, 8727915, perrotta2017application, barbado2021understanding}.

The features are divided into 4 groups: Index, Categorical, Explainable and Target. 

\begin{itemize}
    \item \emph{Index features} refer to features used to identify each row (namely a vehicle's unique id, vehicle\_id, and the date, date\_tx).
    \item \emph{Categorical features} refer to non numerical features used to distinguish group of vehicles (p.e. "vehicle model" indicates vehicles with the same make-model). As mentioned before, they will be covered later, since they are not obtained yet at this point.
    \item Regarding the explainable features, they are further divided into three groups. First, there are features related to the vehicle status itself. For instance, the pressure of the tyres. If the pressure is too low, the fuel needed to cover the same amount of distance will increase, thus increasing the fuel consumption of the vehicle. These features are identified in (Subsection A.2 in the Appendices) as \emph{Vehicle Parameters}. 
    The next group of features are the \emph{Driving Behaviour} ones. They correspond features related to the vehicle's driver behaviour itself that may affect the fuel consumption. An example of these features is the idle time spent. More idle time may increase fuel consumption.
    The last group of features considered are the \emph{Environment Variables}. For instance, the exterior temperature may affect a vehicle's thermodynamic cycle harming its efficiency. 
    \item The final feature is the target column, the fuel consumption itself. This is calculated directly as: 
    \begin{equation}
    fuel\_consumption =  \frac{trip\_fuel\_used}{trip\_kms} \times 100
    \end{equation}
\end{itemize}

This yields a data frame where each row corresponds to the daily aggregated values of the selected features for a specific vehicle. Thus, we want to analyse the potential relationship between those features with the fuel consumption of that vehicle in that day. The data frame generated appears within (Subsection A.3 of the Appendices).

\textbf{Discard target null values}\\
In some cases, the IoT devices may not provide either the fuel spent during that day (trip\_fuel\_used), the distance driven (trip\_kms), or both. Then, for those cases we do not have the value of the fuel consumption. Those records are not considered and are discarded.

\textbf{Eliminate non relevant data}\\
Non-representative vehicle-days are also eliminated, when the distance driven is too low to be significant. A minimum threshold is defined that eliminates all vehicles that have a distance traveled less than that threshold.
Also, this points ensures that there are no highly correlated features within the FAR (though there were none found). We also check if there are no relevant features (none found also). The equations used are shown below, with $x_i$ and $x_j$ being two features.
\begin{itemize}
    \item Remove records with trip distance (Km) $\leq$ min\_day\_km.
    \item Corr($x_i$, $x_j$) $<$ 0.7 \cite{dormann2013collinearity}
    \item Standard Deviation($x_i$) $>$ 0
\end{itemize}

\textbf{Identify vehicle models}\\
The following steps aim to complete the previous features obtained from the IoT devices with relevant categorical features. The categorical features include two different variables. First, a feature named \emph{vehicle\_group}. This features classifies vehicles corresponding to their make-model. Using a vehicle's VIN (Vehicle Identification Number) we can identify their make and model, and group them accordingly. The VIN decoding procedure yields vehicle groups that do also have the same fuel type (diesel or gasoline; our data sets do not include electric or hybrid vehicles).

\textbf{Identify route type}\\
The second feature is \emph{route\_type}. It is used to identify the main type of route of a vehicle in a specific day. We assign a route type for each vehicle-day according to the following rules:

\begin{itemize}
    \item IF $per\_time\_city \leq low\_th\_time$ AND $trip\_kms \geq th\_kms$ THEN $route\_type = hwy$
    \item ELSE IF $per\_time\_city \geq high\_th\_time$ AND $trip\_kms \leq th\_kms$ THEN $route\_type = city$
    \item ELSE $route\_type = combined$
\end{itemize}

Thus, we consider a "city" route if the vehicle spent a minimum amount of time driving within city and if the distance driven does not exceed a specific threshold. On the contrary, to consider the route of a vehicle-day as "highway" (hwy), the time spent driving within city should be lower than a threshold, and the distance driven should be above another threshold. Any other scenario is considered as "combined". This feature is important since the reference fuel consumption of a vehicle is different depending on the route type.

\textbf{Fill missing values}\\ 
In addition to sometimes not having the data related to the target variable, in some cases the IoT devices do not send information about some of the input features. In order to avoid losing excessive registers and maintain a statistically significant set of data, these values are imputed with inferred values from the rest of the fleet.
Separating the data set according to its vehicle\_group, each missing value is assigned with the median value of that feature in order to be able to maintain the record but that the value of that variable for that vehicle-day is not significant to the model. 
This module provides a final data frame ready to be used in the following modules (FAR). The median values considered are from the historical data set. 

\subsection{3.3. Unsupervised anomaly detection}
Using the previous FAR data frame, the next step detects the vehicle-dates where there is an anomalous fuel consumption. Since there is no prior knowledge on when the fuel consumption is anomalous, it needs to detect it in an unsupervised manner. Also, the module needs to provide a threshold value to distinguish outliers from inliers, since we want to include that information as a visual explanation.

To comply with both requirements, we apply an univariate unsupervised anomaly detection approach using a Box-Plot that classifies data points as outliers if they are above or below the thresholds in Equation \ref{eq:boxplot-anomalies}. The Box-Plot will be applied over the different combinations of the categorical variables (make-model with vehicle\_group and route type with route\_type), obtaining then a different limit depending on the combination considered.

\begin{equation}\label{eq:boxplot-anomalies}
\begin{split}
lim\_sup = Q3 + 1.5 \times IQR \\
lim\_inf = Q1 - 1.5 \times IQR
\end{split}
\end{equation}

However, before detecting the fuel anomalies, we use that same Equation to remove vehicles with a fuel consumption that is too small, as well as for removing vehicles with a potential wrong fuel value due to being extremely high. So, we first apply it and remove the vehicles that are above or below the thresholds, and then we apply it again for detecting fuel anomalies. With that, we classify each data point as outlier or inlier.

\subsection{3.4. ML model}
The following module is the training of a ML supervised model that finds relationships between the explainable and categorical features from the FAR dataset and the target variable. 

Since part of this paper is benchmarking several GAM whitebox models, within this step we can use any whitebox model that yields feature relevance-based global explanations. The initial proposal is using EBM \cite{nori2019interpretml}, since its a whitebox algorithm with good predictive power that has been previously used within the fuel consumption context \cite{barbado2021understanding}.

However, there are two problems that arise with EBM within the context of fuel consumption. First, the feature relevance explanations will be exactly the same for all the vehicles within the fleet. That means that the unitary impact from, for instance, one extra speeding event, will be the same for passenger cars than for trucks if the fleet contains both types of vehicles. This could be fixed by using the pairwise terms of EBM in order to adjust each feature. For instance, $f_i(x_i) + f_ij(x_i, x_j)$ will be the adjusted feature relevance value for feature $x_i$ considering the vehicle model $x_j$. The problem is that we will need to adjust every feature combined with every vehicle model, and this significantly increases the number of features used for training the model. Our proposal for this problem is addressed with our EBM variation ("EBM\_var") algorithm, which is detailed later.

Another problem is that the relationship between feature values and feature relevance may not be monotonic when it should be. The original proposal of EBM does not allow for the usage of monotonic constraints. Because of that, we will also evaluate the usage of CGA2M+ algorithm \cite{watanabe2021cga2mplus}, where we can specify monotonic constraints.

\begin{figure}[h!]
\centering
  \begin{tabular}{c@{\qquad}c@{\qquad}c}
  \includegraphics[width=0.85\columnwidth]{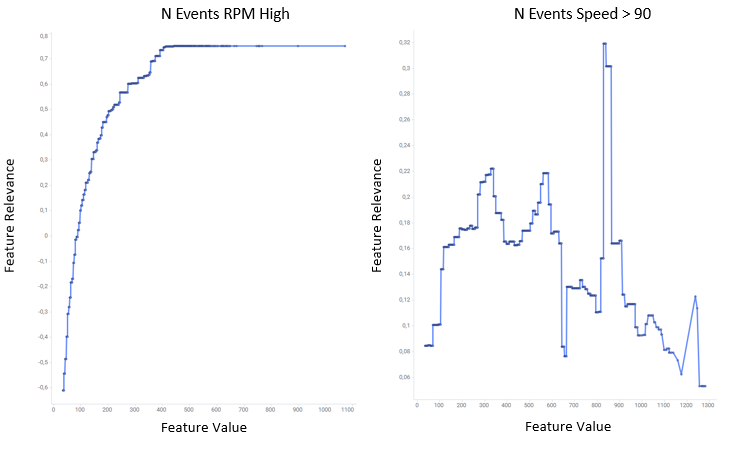}
  \end{tabular} 
  \caption{Problems with EBM. Left, we see that even though the evolution is monotonic by directly using EBM, the model uses one pairplot for every model in the fleet. Right, we see an example of a pairplot that should be monotonic but it is not.}
  \label{fig:ProblemsEBM}
\end{figure}

An example of these problems is shown in Figure \ref{fig:ProblemsEBM}.

Our final solution will use the proposal that yield best results (according to the metrics defined at Subsection 3.8.).

\textbf{EBM variation}\\
The EBM variation that we propose takes into account possible differences that may exist within different subgroups of vehicles, in order to adjust feature relevance and predictions. Regarding our use case, the feature relevance may be different depending on the vehicle group. For instance, the impact on the fuel consumption for each additional harsh brake may change depending on the vehicle's model and make considered. Thus, there should be different feature relevance-values pairs depending on that vehicle group category. Using only one EBM provides unique pairs of value-importance regardless of the vehicle group, meaning that the final impact in the target variable will be the same for a specific feature value.

\begin{figure}[h!]
\centering
 \begin{tabular}{c@{\qquad}c@{\qquad}c}
\includegraphics[width=0.99\columnwidth]{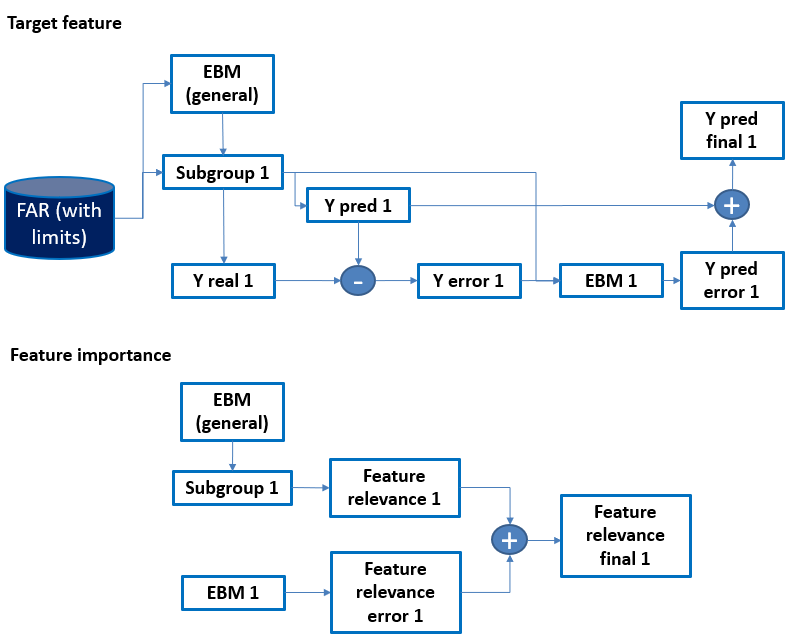}
  \end{tabular} 
  \caption{Proposal of the "EBM variation" over only one subgroup.\label{fig:ebm_variation_flowchart}}
\end{figure}

The intuition behind our proposal is similar to other works in the literature \cite{waeto2017forecasting}. We add an additional layer of models to predict the error of a previous one. As represented in (Figure \ref{fig:ebm_variation_flowchart}) for one subgroup of vehicles, first, we will train an EBM model over all data during the training phase. Then, we will predict the error for each of the vehicle's subgroups, and train additional EBM in order to be able to predict that error and both improve the predictions of the first one as well as adjusting the results to the specificity of each of the subgroups. This last consideration is based on the fact that while the first model provides unique feature relevance-values pairs, because the second one is predicting the error of the first one in order to add it to its prediction, we can also use the feature relevance values of the second one to add them to the first one. This may be done since the feature relevance values of the second model show the feature contribution to the error.
With that, there will be different feature relevance-value pairs, as well as predictions, for each of the vehicle subgroups considered.

The detailed description of EBM variation appears at (Algorithms \ref{alg:ebm_variation_train} and \ref{alg:ebm_variation_exp}). (Algorithm \ref{alg:ebm_variation_train}) describes the training process. The function trainEBMvar receives the input feature matrix $X$ together with the real target variable $y$, and a list with the columns used to consider the subsets, $l_s$. In this case, $l_s$ includes only the variable vehicle\_group. After that, it initializes an empty dictionary $dct\_m$ where the error predicting models are going to be stored. Then, it obtains the potential combination of $l_{comb}$ (in this case, there are no combinations since there is only one feature). Following this, it trains an EBM model using $X$ and $y$.
Iterating through all of the combinations, it filters the input matrix $X$ for the subset for that iteration, $X_i$, getting also the indexes associated to those registers, $idx_i$. If there are not enough data points (less than a threshold $th\_ebm\_var$), it skips that iteration. In other case, it obtains the error for that subset using the original model $emb$, $y\_err_i$. Using that error and the matrix filtered for that iteration, it trains a new model $ebm_i$ that tries to predict the error for that subset. This model is stored within the dictionary $dct\_m$.

After the training, the next step is using those models for prediction and explanations. (Algorithm \ref{alg:ebm_variation_exp}) describes the function expEBMvar used for that purpose. It receives a data frame to explain ($X$), together with the general model ($ebm$), and the dictionary with the models used for error prediction ($dct\_m$). It also receives the list of features for the subsets of data.
The function initialize a data frame to store the feature relevance values ($df\_imp$) and a list with the target feature predictions ($y\_pred$). After obtaining the different combinations for iterating ($l_{comb}$), it firsts predicts the target feature for that subset $X_i$ using the general model $ebm$. Then, if that combination was used for training error predicting models, it obtains the error predictions of the subset, together with their feature relevance values, and adds them to the ones from the original model. If that combination does not belong to any error predicting model, then the function uses only the predictions and feature relevance values from the general model ($ebm$).

\begin{algorithm}[h!]
\caption{EBM Variation training}\label{alg:ebm_variation_train}
\begin{algorithmic}[1]
\Procedure{trainEBMvar}{$X, y, l_s$}
    \State $dct\_m \gets null$
    \State $l_{comb} \gets combinations(X, l_s)$
    \State $ebm \gets trainEBM(X, y)$
    \For{$comb \in l_{comb}$}
        \State $X_i \gets X[X[l_s]=comb]$
        \State $idx_i \gets X_i[index]$
        \If{$len(X_i)<th\_ebm\_var$}
            \State \textbf{continue}
        \EndIf
        \State $y\_pred_i \gets ebm.predict(X_i)$
        \State $y\_real_i \gets y[idx_i]$
        \State $y\_err_i \gets y\_real_i - y\_pred_i$
        \State $ebm_i \gets trainEBM(X_i, y\_err_i)$
        \State $dct\_m[comb] \gets ebm_i$
    \EndFor
    \State \textbf{return} $ebm, dct\_m[comb]$
\EndProcedure
\end{algorithmic}
\end{algorithm}

\begin{algorithm}[h!]
\caption{EBM Variation explanations}\label{alg:ebm_variation_exp}
\begin{algorithmic}[1]
\Procedure{expEBMvar}{$X, ebm, dct\_m, l_s$}
    \State $df\_imp \gets null$
    \State $y\_pred \gets null$
    \State $l_{comb} \gets combinations(X, l_s)$
    \For{$comb \in l_{comb}$}
        \State $X_i \gets X[X[l_s]=comb]$
        \State $y\_pred_i \gets ebm.predict(X_i)$
        \If{$comb$ in $dct\_m$}
            \State $ebm_i \gets dct\_m[comb]$
            \State $y\_err_i \gets ebm_i.predict(X_i)$
            \State $y\_pred_i \gets y\_pred_i + y\_err_i$
            \State $df\_imp\_i \gets ebm.feat\_imp(X_i)$
            \State $df\_imp\_err\_i \gets ebm_i.feat\_imp(X_i)$
            \State $df\_imp\_i \gets df\_imp\_i + df\_imp\_err\_i$
        \EndIf
        \State $y\_pred \gets y\_pred.append(y\_pred_i)$
        \State $df\_imp \gets df\_imp.append(df\_imp\_i)$
    \EndFor
    \State \textbf{return} $y\_pred, df\_imp\_i$
\EndProcedure
\end{algorithmic}
\end{algorithm}

\subsection{3.5. Generate explanations}
Within the "Generate explanations" step, we extract the feature relevance for each data point. This provides a raw data frame with explanations that could be used directly to explain every instance (Subsection A.2 in the Appendices). However, it needs to be combined with business rules in order to select the explanations that comply with them. Since every XAI method considered in this paper establish an additive relationship between input and target features through feature relevance value, from among all available features the process considers only for each data point those that comply with the rules. Generally speaking, for a particular data point \textit{n}, the raw explanations provide the following equations:

\begin{equation}\label{eq:y_pred}
y\_pred (n) = \varepsilon + \sum_{i=1}^{k}  \alpha_{i}(x_{i}(n)) \times x_{i}(n)
\end{equation}

\begin{figure}[h!]
\centering
 \begin{tabular}{c@{\qquad}c@{\qquad}c}
\includegraphics[width=0.8\columnwidth]{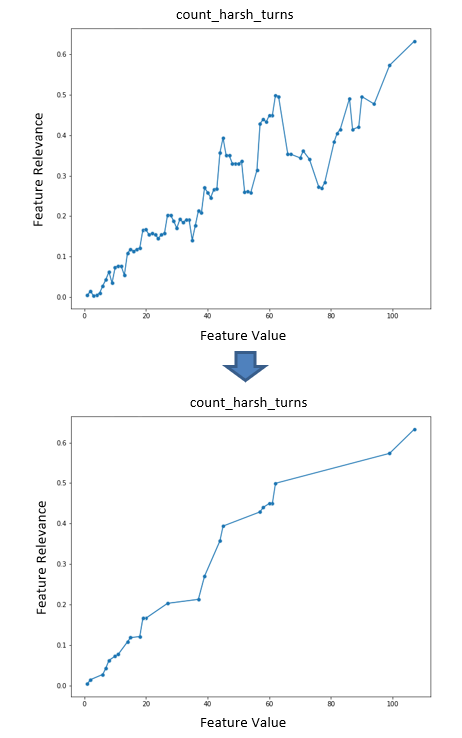}
  \end{tabular} 
  \caption{Example of evolution of the feature value and the feature relevance for feature count\_harsh\_brakes before and after applying the monotonicity filter\label{fig:monotonicity_filter_example}}
\end{figure}

Thus, (Equation \ref{eq:y_pred}) show the relationship for a data point \textit{n} between the predicted value of the target variable y\_pred with respect to \textit{k} input features, $x_{i}$, through their coefficient $\alpha_{i}$. This coefficient $\alpha_{i}$ changes depending on the corresponding $x_{i}(n)$ value for that data point. $\varepsilon$ is a constant intercept term added to the features.
In all the cases, we will train models without using pairwise terms, since they will potentially make the explanations and recommendations too complex. Thus, the explanations will not consider the joint evolution of two features.
\\
\\
\textbf{Business Rules}\\
Over the raw explanations, we apply the following business rules (already introduced in \cite{barbado2021understanding}):
\begin{itemize}
\item BR1: The features used for training the model may be numeric (e.g. time driving uphill) or categorical (e.g. the vehicle model). All those categorical features are one-hot encoded before training the model. However, they are not considered for the explanations since they are not actionable.
\item BR2: We remove the features in the vehicle-date explanations that have a very low impact on the fuel consumption (relative impact below 1\%)
\item BR3: The explanations only include vehicles where the average fuel consumption is above the value of the median inlier vehicles for the same model and on the same route type.
\item BR4: Feature values must be higher than the median value of the vehicle inliers from the same model for that same feature when the feature Type is Positive, or lower when Type is Negative.
\item BR5: The total fuel reduction from the explanations should not be more than the 80\% of the original fuel consumption. Since the models do not allow to impose restrictions in the learning for the individual models for the features, we need to apply this posthoc filtering to remove explanations that are not physically possible.
\end{itemize}
EBM and "EBM\_var" do not necessarily yield monotonic explanations for each feature. Because of that, in this step we include an optional monotonicity filter in order to filter some of the feature relevance - feature values combinations from among all vehicles, and leave only those that result in a monotonic relationship between them. This filtering is optional, and can be used both for selecting only some particular explanations, or for computing a monotonicity metric that measures the degree of monotonicity for each feature in the data set (Subsection 3.8.2). The detailed description of this step is shown below.
\\
\\
\textbf{Monotonicity Filter}\\
The "Monotonicity filter" step analyses each pair of feature value and feature relevance for every vehicle group and route type combination and discards the pairs that are not monotonic. An example can be seen in (Figure \ref{fig:monotonicity_filter_example}). Starting from the evolution of the relevance-value pair of a particular feature, in this step the process finds the feature values intervals where the feature relevance is not monotonic, and discards those combinations. Thus, the raw explanations for each vehicle-day, where all the features are included, are filtered so that the feature values that correspond to feature relevance ones that are not monotonic are now discarded. 
(Figure \ref{fig:monotonicity_filter_example}) shows the original feature relevance-value pairs for a combination of route type and vehicle group for the feature count\_harsh\_brakes. As the Figure shows, the evolution is not monotonic. It also shows the final result after applying the monotonicity filter.

Formally, the step analyses the evolution of the relevance-value pair of every feature for every combination of categorical features as indicated in (Algorithm \ref{alg:monotonicty}). The function "filtMonotonic" receives four variables: $X_{i}$ with the FAR data frame that wants to be explained, $X_{exp}$ with the raw explanations generated previously, $l_e$ with a list of the numerical features (the ones for analysing the monotonicity), and $l_c$ with a list of the categorical columns. Using both $X_{i}$ and $l_{c}$, the function first obtains the possible combination of categorical features and stores that information within $l_{comb}$. Thus, $l_{comb}$ and $l_e$ are the parameters that are going to be considered during each iteration: a unique combination of categorical feature values ($comb$) and one explainable feature ($f$). $comb$ and $f$ are used for filtering the explanations of every vehicle-date of the period in order to have a unique data frame of the importance-value pairs inside that iteration ($X_{check}$). This data frame is sorted in an ascending order using the feature value.
After that, the function gets the difference of the feature relevance between one feature value and the following one. If the evolution is monotonic, the difference should be 0 or higher (0 because we only check for monotonic evolution, not strictly monotonic). The function discards the rows that are not monotonic, and keeps checking the difference of feature relevance between one row and the following one until no rows are discarded (which means that the data frame is already monotonic).

\begin{algorithm}[h!]
\caption{Monotonicity}\label{alg:monotonicty}
\begin{algorithmic}[1]
\Procedure{filtMonotonic}{$X_{i}, X_{exp}, l_e, l_c$}
    \State $X_{exp\_new} \gets null$
    \State $l_{comb} \gets combinations(X_{i}, l_c)$
    \For{$comb \in l_{comb}$}
        \For{$f \in l_{n}$}
            \State $X_{check} \gets filter(X_{exp}, comb, f)$
            \State $X_{check} \gets dropDuplicates(X_{check})$
            \State $X_{check} \gets sort(X_{check})$
            \State $n_{diff} \gets -1$
            \While{$n_{diff} \neq 0$}
                \State $n_{i} \gets len(X_{check})$
                \State $X_{check}['diff'] \gets getDiff(X_{check})$
                \State $X_{check} \gets X_{check}['diff'] \geq 0$
                \State $n_{e} \gets len(X_{check})$
                \State $n_{diff} \gets n_{i} - n_{e}$
            \EndWhile
            \State $X_{exp\_new} \gets append(X_{exp\_new}, X_{check})$
        \EndFor
    \EndFor
    \State \textbf{return} $X_{exp\_new}$
\EndProcedure
\end{algorithmic}
\end{algorithm}

Since the monotonicity filter analyses the combined evolution of both feature relevance and feature value, it works either for EBM (where there is only one value-importance pair per feature at the dependency function \cite{nori2019interpretml}), "EBM variation" (where there is potentially one value-importance pair per feature and vehicle group), or LIME and SHAP (where there may be more than one importance value per unique feature value \cite{molnar2019interpretable}). Indeed, there may be more than one importance-value pair per feature value. However, since (Algorithm \ref{alg:monotonicty}) checks a pair and the immediate following one, it will, for instance, check $(x0,y0)$ against $(x0, y1)$ with $y1 > y0$, and will remove the latter if the importance is lower.

A final comment is that, in some cases, there may be only one feature relevance-value pair because there was only one instance to begin with, or there is only one remaining instance after applying the filter. In those cases, the instance is kept.

With that, the output for this step includes the feature relevance (after filtering with business rules, and optionally with the monotonicity filter) for each of the anomalous vehicles and dates included within the date range considered for the explanation phase. 

\subsection{3.6. Daily recommendations}
Whitebox models that include feature relevance are useful for counterfactual explanations. Since there is a unique intercept and unique feature relevance-value pairs, they can provide counterfactual explanations where one of the feature values alone may be changed and with that, recalculate the predicted target value in order to see how it will change. These counterfactual explanations are, in fact, a recommendation, since they the show future scenarios if particular actions take place.

The intuition behind it is the following one. "Generate Recom." changes the feature values of the outliers used within the explaining phase for the corresponding median feature value of the inliers belonging to the same vehicle group and route type. This will be applied for one feature at a time and for every feature labeled as "actionable". Then, by substracting the relative change in the predicted value from the real fuel consumption, it will indicate which vehicles-dates would have a fuel consumption below the outlier limit for that vehicle group and route type. 

The details are described in (Algorithm \ref{alg:get_recom}); getRecom function receives the historical median values of the inliers (obtained during the training phase; $X_{med}$), the data points of the explaining phase with their feature relevance ($X_{exp}$), and two lists, one with the explainable features that are actionable ($l_a$) and one with the categorical ones ($l_c$). It also receives a list l\_z with the features that going to be explained using a zero reference. For instance, by reducing the "harsh brakes" to zero, instead of the median value for that vehicle group. Using these inputs, getRecom function initializes two empty lists ($l\_up\_ind$ and $l\_up\_all$) and gets the feature relevance for the median inliers feature values ("coeff"), or zero value, with $checkPairwise(X_{med}, l_c, l_z)$ function. 
After obtaining the feature relevances, the function analyses every data point ($x$) within the explanations and obtains it's predicted target feature ($y\_pred$) using the feature relevance and the intercept. It also stores the real value ($y\_real$) of the target feature. 
Then, it checks every feature ($f$) within the explanations and gets its corresponding feature relevance from the median inliers reference, or the zero reference, ($\beta_{fn}$). It laters sums again all the feature relevance and intercept for data point x, without the feature relevance for feature "f", and instead sums $\beta_{fn}$. This leads to a new predicted value ($y\_new$) where all the other feature values are kept the same, but there is a change for the specific feature considered. The difference between $y\_pred$ and $y\_new$ is $\Delta$, and this difference is used to compute the change in the real fuel consumption ($l\_up\_ind$). After iterating for all the available combinations, getRecom uses groupVal function to obtain the estimated value in case all the actionable features change at the same time to their median inlier value, or to the zero value. This is simply done by aggregating all the individual changes in the prediction for each feature, and subtract the aggregated difference from the real fuel consumption.

\begin{algorithm}[h!]
\caption{Generate Recommendations}\label{alg:get_recom}
\begin{algorithmic}[1]
\Procedure{getRecom}{$X_{med}, X_{exp}, l_a, l_c, l_z$}
    \State $l\_up\_ind \gets null$
    \State $l\_up\_all \gets null$
    \State $coeff \gets checkPairwise(X_{med}, l_c, l_z)$
    \For{$x \in X_{exp}$}
        \State $y\_pred \gets \varepsilon + \sum_{i=1}^{k}  F_{i}(x_{i}) $
        \State $y\_real \gets x[target]$
        \State $comb \gets x[l_c]$
        \For{$f \in l_a$}
            \State $\beta_{fn} \gets coeff[f]$
            \State $y\_new \gets \varepsilon + \sum_{i=1}^{k \neq f}  F_{i}(x_{i})$
            \State $y\_new \gets y\_new + \beta_{fn}$
            \State $\Delta \gets y\_pred - y\_new$
            \State $y\_updated \gets y\_real - \Delta$
            \State $l\_up\_ind \gets l\_up\_ind.append(y\_updated)$
        \EndFor
    \EndFor
    \State $l\_up\_group \gets groupVal(l\_up\_ind, l_a, X_{exp}, l_c)$
    \State \textbf{return} $l\_up\_ind, l\_up\_group$
\EndProcedure
\end{algorithmic}
\end{algorithm}

Thus, (Algorithm \ref{alg:get_recom}) provides a list with the new estimated fuel consumption value for every individual feature change for every vehicle-date pair ($l\_up\_ind$). Comparing this values against the outlier limit for that vehicle group and route type, the step indicates which individual feature changes will lead from outlier to inlier, and what would be the corresponding fuel consumption. It provides as well a similar result but considering that every actionable feature changes at the same time ($l\_up\_group$). 

\subsection{3.7. Recommendations according to user profiles}
According to \cite{alej2019explainable} explanations should be tailored for the specific profile of the user that will receive them taking into account both their expectations and their domain knowledge. 

\begin{figure}[h!]
\centering
 \begin{tabular}{c@{\qquad}c@{\qquad}c}
\includegraphics[width=1\columnwidth]{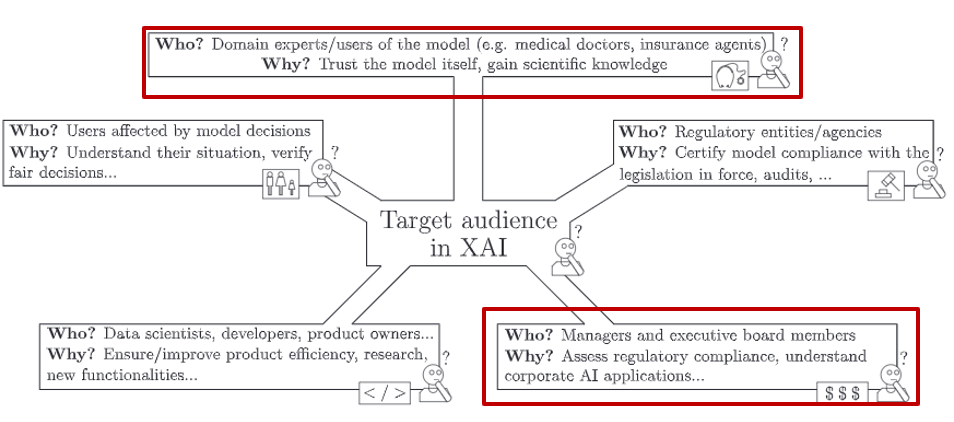}
  \end{tabular} 
  \caption{Relevant user profiles for this use case, following the proposal of \cite{alej2019explainable}}
  \label{fig:user_profiles}
\end{figure}

Within the use case proposed in this paper, we identify two user's profiles, as indicated in (Figure \ref{fig:user_profiles}), where the users are highlighted over the image from \cite{alej2019explainable}. 

\subsubsection{User Profile 1: Technical Specialists}
The first group of users are technical specialists, responsible for the status of the vehicles. Their main interest in explanations is detecting what vehicles are consuming excessively, and what is causing it, considering for that not every feature, but only the ones that are actionable. This corresponds to all the features in (Subsection A.2 in the Appendices) with the exception of "trip kms" and "odometer".

To accomplish that, the recommendations generated at (Section 3.6) may be enough. However, providing information every single date for each combination of vehicles and route types in terms of the numeric feature relevance is overwhelming, not being useful for them. This is why we provide the recommendations for these users at two different levels. First, a summary of the main recommendations for a specific period of time (p.e. a month). Second, we provide the individual daily detail only if the want to dive deeper into a particular vehicle and route type. In both cases, we only include vehicles with fuel consumption anomalies.

\textbf{First level - Summary of recommendations}\\
The first level includes a summary of the individual recommendations yielded by the system. (Algorithm \ref{alg:get_recom_summ}) described the way to accomplish it. First, it receives the same input, $X_{med}, X_{exp}, l_a, l_c$, as (Algorithm \ref{alg:get_recom}). The difference is that, before obtaining the recommendations, it applies a filter that chooses only some vehicles and route types, from among all the combinations, according to some business parameters. These parameters are $min\_days\_anomalies$, $min\_day\_km$, and $min\_dev\_total\_avg\_fuel$. With $min\_days\_anomalies$, the filter chooses only the vehicle-route type combinations that have at least that specified number of outliers. Then, with $min\_day\_km$, it chooses only the dates that have a trip distance over that minimum threshold. Finally, with $min\_dev\_total\_avg\_fuel$, the filter chooses only dates with individual recommendations that have a decrease in the target variable after applying the recommendations over that threshold.

After applying the aforementioned filters with $filterPoints()$ function, the algorithm applies another function, $summaryPoints()$. This function aggregates the remaining individual data points of the outliers into their median values. So, it will yield a data frame with unique points for each combinations of vehicles-route type. These points will represent a prototype for each of those combinations, representing the most common anomalous scenario. These data points are stored in $X_{summ}$. In order to always have feature values already present within the explanation period, if the vehicle values are pairs (not odds) we keep the lowest middle value in order to offer later the most conservative recommendation.
Then, the algorithm uses $X_{summ}$ for obtaining the recommendations with $getRecom$ function. In this case, we are only interested in the output $l\_up\_ind$, that indicates the new fuel consumption after applying each individual feature change in order to have the median inlier value.

These individual contributions are aggregated with $aggContribution()$ function, providing $l\_agg$ with the total fuel consumption reduction if all the features had the value of the median of the inliers of that same vehicle group and route type. 
With that, the user will see the general recommendations (how much average fuel could be decreased by applying all the feature value changes), as well as seeing the individual impact of each feature to the fuel consumption (seeing how much fuel consumption could be reduced by applying only one feature change).

\begin{algorithm}[h!]
\caption{Summary of Recommendations}\label{alg:get_recom_summ}
\begin{algorithmic}[1]
\Procedure{getSummRecom}{$X_{med}, X_{exp}, l_a, l_c, l\_z$}
    \State $X_{exp} \gets filterPoints(X_{exp})$
    \State $X_{summ} \gets summaryPoints(X_{exp})$
    \State $l\_up\_ind, _ \gets getRecom(X_{med}, X_{summ}, l_a, l_c, l\_z)$
    \State $l\_agg \gets aggContribution(X_{summ}, l\_up\_ind)$
    \State \textbf{return} $l\_up\_ind, l\_agg$
\EndProcedure
\end{algorithmic}
\end{algorithm}

\begin{figure}[h!]
\centering
 \begin{tabular}{c@{\qquad}c@{\qquad}c}
\includegraphics[width=0.95\columnwidth]{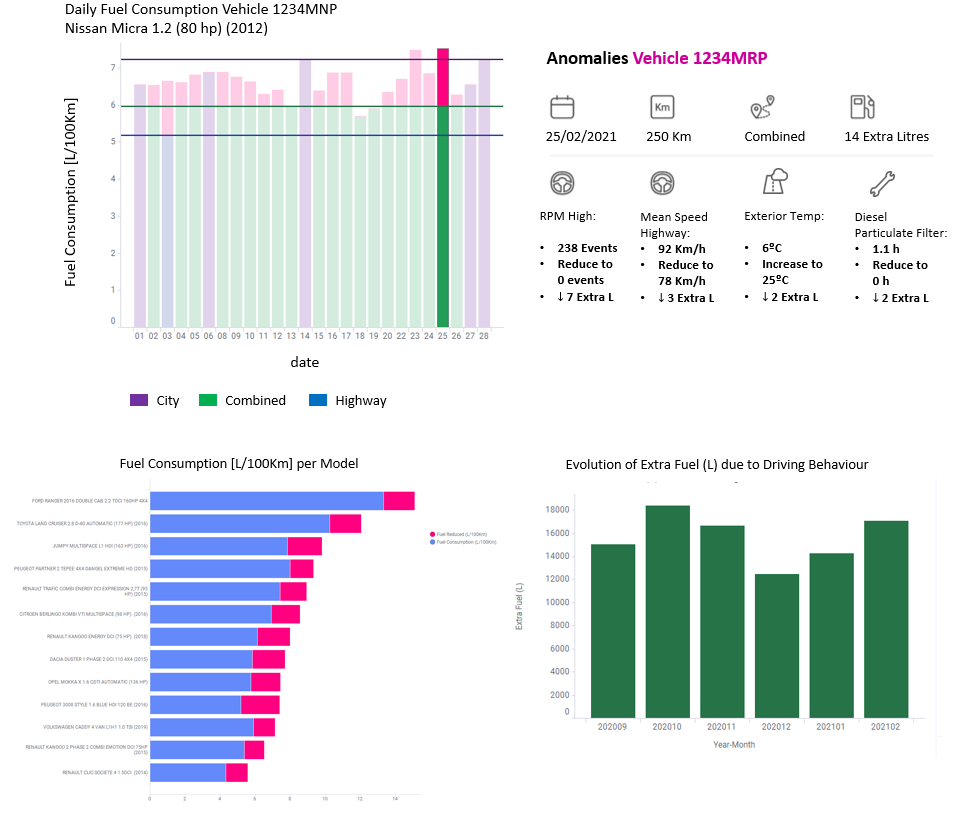}
  \end{tabular} 
  \caption{Example of explanations and recommendations for Fleet Operators (above) and Fleet Managers (below).\label{fig:ExplanationsDashboardExample}}
\end{figure}

\textbf{Second level - Daily detail}\\
If the user wants to see recommendations for individual days, they can access to these second explainability level. It includes both the recommendations from (Subsection 3.5), along with the visual explanations for the anomalies using the upper limits from (Subsection 3.3). An example of this output is included in Figure \ref{fig:ExplanationsDashboardExample}.

\subsubsection{User Profile 2: Fleet Manager}
The final user profile considered is the "fleet manager". The main interest for this user profile is having a global comparative view at a vehicle model level, not seeing information about individual vehicles or particular dates. The useful explanations should be expressed in terms of extra litres of fuel consumed, because that can be immediately turned into an economic cost, as well as in an environmental impact. It is also specially useful for this profile not to consider all types of features in the explanations, but only the ones related to driving behaviour, since they are among the features with more impact \cite{zacharof2016review}, they are actionable, and they are mainly associated to inefficient driving styles.
With that, after having the individual recommendations from (Algorithm \ref{alg:get_recom}), the individual explanations are aggregated for the whole fleet and for each vehicle model, considering only for the potential fuel reduction features related to driving behaviour. A final comment is that these explanations include all data points, not only the outliers, since it is an aggregated view.

\subsection{3.8. Metrics}
There are two aspects to measure through metrics: model performance and quality of the explanations/recommendations. For the first case, we measure the predictive power of the ML surrogate model by seeing how close are the predictions to the real fuel consumption value of the different vehicle-dates. We will further refer to them as \textit{model metrics}. 

The second group of metrics are the ones obtained during the explaining phase, and they aim to measure different aspects regarding the understandability of the explanations generated, that can be useful for analysing the explanations by themselves, as well as for comparing the explanations generated between every model. We will further refer to them as \textit{XAI metrics}. 

With that, we are analysing not only if the predictive power of the interpretable models are good enough (model metrics), but we are also measuring the explanations themselves in order to compare them between them and by themselves.
A final comment is that even tough it may be difficult (or not reliable) to compare individual explanations with certain metrics due to Rashomon's Effect \cite{molnar2019interpretable}, the ones that we propose analyse the explanations from a general perspective. Thus, even if the explanations differ at a very low level, the general view should be similar.

\subsubsection{3.8.1 Model metrics}
\leavevmode\newline
Model metrics first include metrics used for comparing the models among themselves. Here we use a test set in order to evaluate the model performance metrics. For that, we use we use the Adjusted R2 value (adj-R2) and the Mean Average Percentage Error (MAPE). 

All the model metrics are evaluated over a test set that includes both outliers and inliers, since the purpose is to measure how close are the target feature predictions to the real value. There are other potential metrics that can be considered, especially classification metrics that measure if after applying the anomaly limits over the predicted values, the inlier/outlier predicted class matches the one of the real target feature. However, since we are not using the ML surrogate model to actually predict the outlier/inlier class, we did not use them. 

\subsubsection{3.8.2 XAI metrics}
\leavevmode\newline
Using the taxonomy of metrics in \cite{carvalho2019machine} for individual explanations, we consider different properties for comparing the explanations generated by the different methods studied in this paper. The properties considered are \textit{representativeness}, \textit{precision}, \textit{stability}, \textit{contrastiveness} and \textit{consistency with apriori beliefs}, since they address the main aspects of our use case. For stability and precision, we use metrics based on already existing metrics within the literature. For representativeness, contrastiveness and consistency with apriori beliefs, we propose additional ones that are useful for benchmarking the models within our use case. 
These metrics are used for evaluating the final explanations provided by the system (after applying all the business rules mentioned in the previous subsections), and they are evaluated for anomalous vehicle-dates only. A summary of these metrics appear in Table \ref{table:xai_resume_metrics}.

\textbf{Representativeness} metrics aim to measure the relevance or importance of the explanation. They include these metrics:
\begin{itemize}
    \item \textbf{n\_features}: Number of features used for fuel explanations for a particular vehicle-date.
    \item \textbf{rel\_importance}: Percentage of fuel covered by the features used in the explanations.
\end{itemize}

\textbf{Precision} metrics measure how close is the fuel prediction to the real fuel value using the features within the explanations. For that, we use the MAPE obtained by comparing that fuel prediction based on the final explanations against the real fuel consumption value.
These metrics appear in Equation \ref{eq:xai-metrics-eq-1}, where $v$ is one vehicle, $t$ one date, $X^{'}(v,t)$ the remaining features after applying the business rules, $f_{i}$ a particular dependency function for feature $i$, $y\_p^{'}(v, t)$ the fuel prediction using the features remaining after the business rules, $card$ the cardinality, $N_t$ the number of data points for that vehicle $v$, and $y\_r(v,t)$ the real fuel value for that vehicle and date.

\begin{equation}\label{eq:xai-metrics-eq-1}
\begin{aligned}
n\_features_{v,t} = card(X^{'}(v,t))
\\
rel\_importance_{v,t} = \frac{\beta_0 + \sum_{i=0}^{n\_features_{v,t}} {\beta_i \times f_{i}(x_{v,t})}} {y\_r(v,t)}
\\
xai\_mape_{v} =  \frac{\sum_{t=0}^{N_t} {mape(y\_p^{'}(v, t), y\_r(v,t))} } {N_t}
\end{aligned}
\end{equation}

\textbf{Stability} metrics includes one metric, \textit{stability\_error}. This metric is based on the proposal of \cite{melis2018towards}, as indicated in (Equation \ref{eq:stability_metric_sample}). It computes the norm of the difference for the explanation-based predictions for the two closest data points within the data set. These two data points are found by using a K-Nearest Neighbours algorithm over the same input features used for training the ML surrogate model. This value is then scaled considering the distance between those two data points (in order to penalize the metric if they are not very close). The formula appears in Equation \ref{eq:stability_metric_sample}, where $x_i$ and $x_j$ are two data points, $f_{expl}$ the predictions based explanations for those data points, and $h$ the distance between them. For this particular metric, we do not filter the explanations using the business rules.

\begin{equation}\label{eq:stability_metric_sample}
\qquad \includegraphics[width=0.8\columnwidth, valign=c]{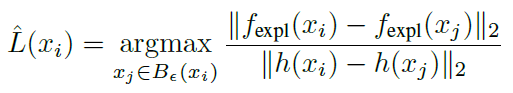}
\end{equation}

\textbf{Contrastiveness} metrics measure the impact of the recommendations generated over the explanations provided by the XAI technique. Because they focus on the recommendations, for our use case they are only applicable to whitebox models. These metrics are:
\begin{itemize}
    \item \textbf{per\_var}: Percentage of fuel saved for a particular vehicle-date after applying the recommendations provided by the system.
    \item \textbf{per\_below}: Percentage of anomalous vehicle-dates, within each vehicle model, that receive fuel recommendations that would change their fuel value below the anomaly threshold value.
\end{itemize}

\textbf{Consistent with Apriori Beliefs} metrics measure how aligned are the explanations with prior domain knowledge. It includes the following metrics:
\begin{itemize}
    \item \textbf{per\_mon}: Percentage of data points for a particular feature and model that are monotonic after applying Algorithm \ref{alg:monotonicty}.
    \item \textbf{MAPE vs reference fuel}: This metric computes the MAPE value of the new average fuel consumption (after the recommendations) against the catalog fuel reference for that model. This metric appears in \cite{barbado2021understanding}.
    \item \textbf{\% below catalog}: It shows the percentage of vehicle-dates that are receiving a recommendation that turns the average fuel consumption (L/100Km) below the catalog reference (with an offset of 1 L/100Km). It should be minimized, because the target fuel should not be below the catalog reference (is a value that is not physically reachable). This metric appears in \cite{barbado2021understanding}.
\end{itemize}

These metrics appear in Equation \ref{eq:xai-metrics-eq-2}, using the results from previous Algorithms \ref{alg:monotonicty} and \ref{alg:get_recom}, and with $N_{m,v,t}$ the vehicle-dates for model $m$.

\begin{equation}\label{eq:xai-metrics-eq-2}
\begin{aligned}
per\_var_{v,t} = \frac{getRecom({X_{med}, x(v,t), l_a, l_c, l_z})[0][0]} {y\_r(v,t)}
\\
per\_below_{m} = \frac{\sum_{t=0}^{N_{m,v,t}} x_n(v,t) } {N_{m,v,t}}
\\
\\
x_n(v,t) =
\\
    \begin{cases}
      1 & getRecom({X_{med}, x(v,t), l_a, l_c, l_z})[0][0] \leq l(m,r)\\
      0 & \text{otherwise}
    \end{cases}     
\\
per\_mon = \frac{card(filtMonotonic(X_{i}, X_{exp}, l_e, l_c)[f, m])} {X_{exp}[f, m]}
\\
\end{aligned}
\end{equation}

\begin{table}[h!]
\centering
\resizebox{130pt}{!}{%
\begin{tabular}{@{}lll@{}}
\toprule
\textbf{Taxonomy}  & \textbf{Metric}       \\ \midrule
Representativeness & n\_features           \\
Representativeness & rel\_importance  \\
Precision          & xai\_mape             \\
Stability          & stability\_error      \\
Contrastiveness    & per\_var              \\
Contrastiveness    & per\_below            \\
Apriori Beliefs    & per\_mon     \\
\bottomrule
\end{tabular}%
}
\caption{Summary of the XAI metrics analysed, linking them to their taxonomy.}
\label{table:xai_resume_metrics}
\end{table}

\section{4. Evaluation}
We use our algorithm over different IoT real-worlds industry data sets, to evaluate the following hypotheses:

Regarding the "Model Evaluation" step:
\begin{itemize}
    \item \textbf{H1}: EBM\_variation and CGA2M+ models yield good enough metrics regarding model's performance. 
\end{itemize}
Regarding the "XAI Evaluation" step:
\begin{itemize}
    \item \textbf{H2}: EBM\_variation and CGA2M+ models yield at least similar XAI metrics for the explanations of anomalous data points (in terms of representativeness, precision, stability, contrastiveness and consistency with a priory beliefs) when compared to EBM.
\end{itemize}

\subsection{4.1. Data sets}
We consider 9 data sets, belonging to different fleets, as appears in (Table \ref{table:data-description}). These data sets are samples for some of their vehicles, and the aggregated information includes information collected during 2019, 2020 and 2021. The table indicates the data set (Fleet), the number of individual vehicles (Vehicles), the number of vehicle groups (Models), the unique combinations of vehicle-dates (Points), the N points that are associated with an anomalous fuel consumption according to the proposal of Section 3 (Outliers), and how many of those data points are within the test set (Outliers [test]). Together with that, we also include a fleet size category (Fleet Size) following the one that appears in \parencite{fleet2021trends}, where fleets with more than 500 vehicles are considered "enterprise) (or large), fleets between 50 and 499 "medium", and fleets with less than 49 vehicles "small". ALl the vehicles have either petrol or diesel engines.

\begin{table}[h!]
\centering
\resizebox{230pt}{!}{%
\begin{tabular}{@{}lllllll@{}}
\toprule
\textbf{Fleet} &
  \textbf{Vehicles} &
  \thead{\textbf{Fleet} \\ \textbf{Size}}&
  \textbf{Models} &
  \textbf{Points} &
  \textbf{Outliers} &
  \thead{\textbf{Outliers} \\ \textbf{[test]}}
  \\ \midrule
D1 & 1552 & Large & 16 & 219707 & 5772  & 577    \\
D2 & 1568 & Large & 16 & 121160 & 1809  & 181    \\
D3 & 316  & Medium & 44 & 65549  & 10484 & 1046   \\
D4 & 252  & Medium & 14 & 35394  & 1944  & 193    \\
D5 & 165  & Medium & 20 & 22478  & 724   & 71     \\
D6 & 143  & Medium & 20 & 18635  & 2003  & 201    \\
D7 & 33   & Small & 5  & 9733   & 949   & 95     \\
D8 & 20   & Small & 5  & 2235   & 349   & 35     \\
D9 & 3    & Small & 2  & 300    & 10    & 2      \\ \bottomrule
\end{tabular}%
}
\caption{Dataset description, including the number of datapoints, number and type of vehicles.}
\label{table:data-description}
\end{table}

A model is trained over each one of those data sets, using the 90\% of the data points for training, and the remaining 10\% for testing. As already mentioned, the model's performance metrics are analyzed considering all the test data points (comparing the model's prediction of the average fuel consumption versus the real value). For the XAI metrics used for comparing the explanations generated, we use the outlier data points that are within the test set (the ones that appear within the explanations and recommendations from the method in Section 3).

Regarding the business variables described in (Section 3.5), after a validation with domain experts, the values chosen for the evaluations conducted are the following ones:

\begin{itemize}
    \item $th\_kms = 30$
    \item $low\_th\_time = 0.5$
    \item $high\_th\_time = 0.65$
    \item $th\_ebm\_var = 100$
    \item $min\_day\_km = 5$
    \item $min\_days\_anomalies = 3$
    \item $min\_dev\_total\_avg\_fuel = 1$
\end{itemize}

\subsection{4.2. Model configuration}
The hyperparameters used for every model match the default ones provided by the software libraries used (only modifying the parameters related to the monotonic constraints) since we did not find any significant improvements after using a grid search over the training data. Regarding "EBM variation", both the general EBM and the EBM for error prediction within the different subgroups use the same hyperparameter configuration.

\subsection{4.3 Model evaluation}
For checking \textbf{H1}, we first analyse the results from EBM\_var and CGA2M+ over the different data sets in order to see if the results are good enough. As already mentioned, we use adjusted R2 and MAPE, since they both yield a result in terms of percentage that can be easily understood.

Regarding adjusted R2, even if it's clear that it indicates the proportion of the variance in the target feature that can be predicted using the input features, it is not trivial to define value thresholds to indicate if the model is good or not. It heavily depends on both the context and the units of the target feature \cite{hair2016primer, hair2013partial}. However, there are some guidelines that may be considered. As a reference, we use the proposal of \cite{chin1999structural} that mentions the following levels:
\begin{itemize}
    \item $0.67$: Substantial
    \item $0.33$: Moderate
    \item $0.19$: Weak
\end{itemize}

MAPE, is a metric commonly used for forecasting models. However, it can be also useful for regression tasks \cite{de2016mean}. Though it is also not direct to define thresholds for MAPE, we use as reference the ones detailed in \cite{lewis1982industrial}, originally proposed for forecasting models.

\begin{itemize}
    \item $<0.1  $: Highly accurate forecasting
    \item $0.1-0.2$: Good forecasting
    \item $0.2-0.5$: Reasonable forecasting
    \item $>0.5$:   Inaccurate forecasting
\end{itemize}

The metrics over the test set for each ML model used for predicting the fuel consumption over each one of the data sets considered appear in Table \ref{table:mape-r2-test}, where we see the mean MAPE over each vehicle in every data set, as well as the adjusted R2 metric for all the predictions in every data set. For EBM\_var, we  see that in all data sets the MAPE value is "highly accurate", with the exceptions of D3, D4 and D9) where is "good". The same happens with CGA2M+, with the exception of D4, which is "highly accurate" in this case . For Adjusted R2, EBM\_var is always within the "substantial" category, except for the case of D4, where it is "moderate". 

\begin{table}[h!]
\centering
\resizebox{225pt}{!}{%
\begin{tabular}{@{}lllllll@{}}
\toprule
\textbf{\begin{tabular}[c]{@{}l@{}}Data \\ set\end{tabular}} &
  \textbf{\begin{tabular}[c]{@{}l@{}}MAPE \\ EBM\end{tabular}} &
  \textbf{\begin{tabular}[c]{@{}l@{}}MAPE\\ EBM\_var\end{tabular}} &
  \textbf{\begin{tabular}[c]{@{}l@{}}MAPE\\ CGA2M+\end{tabular}} &
  \textbf{\begin{tabular}[c]{@{}l@{}}Adjst R2\\ EBM\end{tabular}} &
  \textbf{\begin{tabular}[c]{@{}l@{}}Adjst R2\\ EBM\_var\end{tabular}} &
  \textbf{\begin{tabular}[c]{@{}l@{}}Adjst R2\\ CGA2M+\end{tabular}} \\ \midrule
D1 & 0.08 & 0.08 & 0.08 & 0.77 & 0.8  & 0.79 \\
D2 & 0.09 & 0.08 & 0.08 & 0.66 & 0.84 & 0.85 \\
D3 & 0.13 & 0.11 & 0.14 & 0.94 & 0.96 & 0.92 \\
D4 & 0.09 & 0.10 & 0.09 & 0.61 & 0.64 & 0.61 \\
D5 & 0.09 & 0.08 & 0.08 & 0.66 & 0.72 & 0.72 \\
D6 & 0.09 & 0.07 & 0.08 & 0.85 & 0.9  & 0.86 \\
D7 & 0.08 & 0.07 & 0.09 & 0.8  & 0.83 & 0.82 \\
D8 & 0.08 & 0.08 & 0.06 & 0.63 & 0.69 & 0.67 \\
D9 & 0.15 & 0.15 & 0.17 & 0.85 & 0.84 & 0.81 \\ \bottomrule
\end{tabular}%
}
\caption{MAPE and Adjusted R2 over the test set for each ML model and for each data set.}
\label{table:mape-r2-test}
\end{table}


\subsection{4.4 XAI evaluation}
\textbf{H2} is evaluated by analysing several XAI metrics in terms of representativeness, precision, stability, contrastiveness and consistency with apriori beliefs. The results are shown in Table \ref{table:xai-metrics-contrast}, showing the Kruskal-Wallis hypothesis contrast test \cite{kruskal1952use} comparing the results between the three interpretable methods.
Considering \textit{representativeness} metrics, we see how there are significant differences in all the comparisons. For the number of features used, "EBM\_var" outperforms EBM in every type of data set, while being also outperformed by CGA2M+. Regarding the relative importance, for Large data sets, EBM improves the other two methods (thus, it uses less features but covers more fuel). In Medium and Small data sets the best results are for CGA2M+.

For \textit{precision}, we see less cases where there are significant differences. In fact, only in Medium data sets we see how EBM and "EBM\_var" outperform CGA2M+. 

Regarding \textit{contrastiveness}, there are again many cases where the differences are statistially significant. In Large data sets, "EBM\_var" yields the best results in terms of percentage of fuel variation from the daily recommendations. Therefore, in Medium and Small data sets, the best results are provided by CGA2M+. For the percentage of data points that will be below the anomaly threshold, we see that the results are generally better for "EBM\_var" since there are significant differences where the metric is better for Large and Medium data sets. In small data sets, however, CGA2M+ is better, covering almost all the fuel anomalies.

In the case of apriori beliefs, we see the degree of monotonicity for EBM and "EBM\_var", compared to the perfect degree of monotonicity from CGA2M+. We see that in both cases the degree of monotonicity is significantly lower than the perfect score, and that EBM is significantly more stable than "EBM\_var" in terms of this metric.

Continuing with the measurements of apriori beliefs, we check the MAPE value of the new average fuel consumption for each vehicle-date after applying the recommendations, compared to the catalog fuel consumption for the same make, model, year, fuel type and route type. For this analysis we only use D1, since it is the largest fleet and because we know exactly the vehicle models and its associated catalog fuel consumption reference. Results appear in Table \ref{table:comparison-fuel-reference}, with MAPE 1 corresponding to the median MAPE versus the catalog fuel, MAPE 2 corresponds to the median MAPE versus the median fuel inlier vehicles, and MAPE 3 is the same as MAPE 2 but considering only explanations for outlier vehicles. The results are similar for the three models, with CGA2M+ being better for the percentage of vehicles below the catalog fuel reference. 

\begin{table}[h!]
\centering
\resizebox{190pt}{!}{%
\begin{tabular}{@{}llllllll@{}}
  \textbf{Method} &
  \rot{\textbf{MAPE 1}} &
  \rot{\textbf{MAPE 2}} &
  \rot{\textbf{MAPE 3}} &
  \rot{\textbf{\% MAPE 1 \textless 0.5}} &
  \rot{\textbf{\% MAPE 2 \textless 0.2}} &
  \rot{\textbf{\% MAPE 1 \textless 0.1}} &
  \rot{\textbf{\% Below Catalog}} \\ \midrule
EBM      & 0.14 & 0.14 & 0.14 & 94.2 & 64.2 & 36.8 & 3.3 \\
EBM\_var & 0.15 & 0.14 & 0.14 & 93.2 & 62.6 & 35.6 & 3   \\
CGA2M+    & 0.14 & 0.13 & 0.14 & 94.4 & 63.9 & 36.7 & 2.6 \\ \bottomrule
\end{tabular}%
}
\caption{Different MAPE metrics for each of the models versus the catalog fuel consumption (MAPE 1), the median inliers (MAPE 2), or considering only the vehicles with outlier fuel consumption versus the inliers (MAPE 3).}
\label{table:comparison-fuel-reference}
\end{table}

\begin{figure}[h!]
\centering
 \begin{tabular}{c@{\qquad}c@{\qquad}c}
\includegraphics[width=0.9\columnwidth]{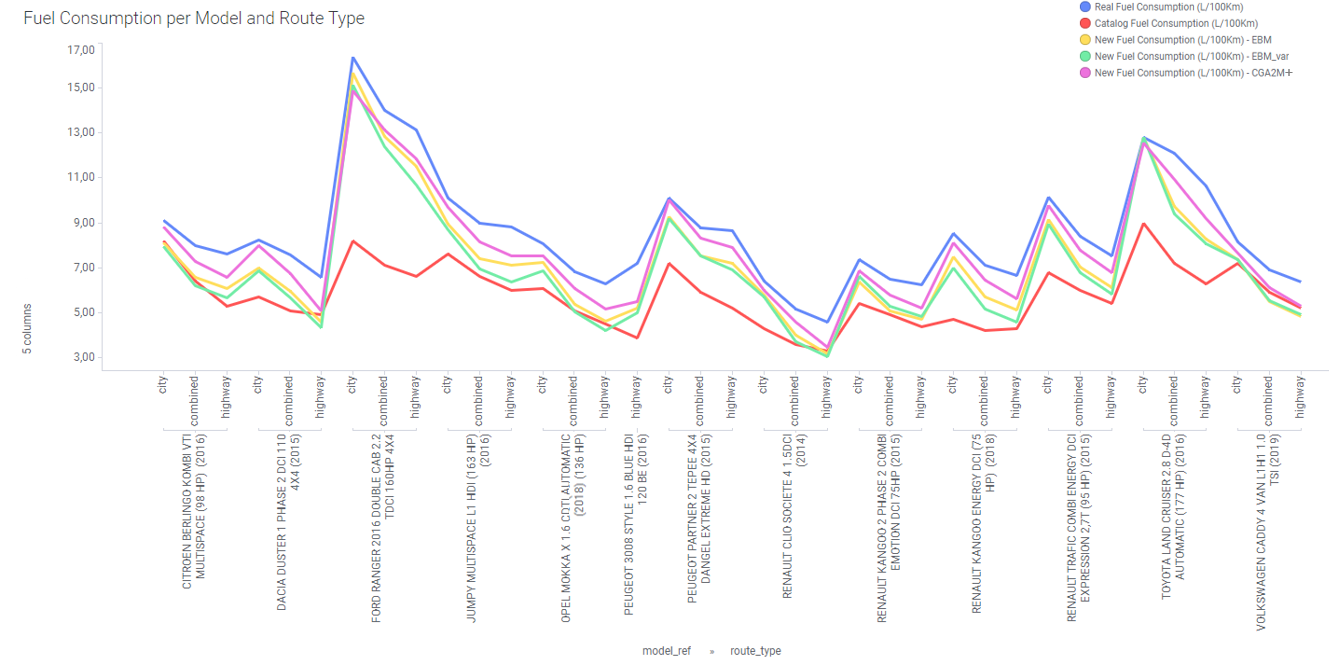}
  \end{tabular} 
  \caption{Comparison of the potential fuel reduction per vehicle model and route type for D1. The comparison includes the three algorithms with respect to both the real fuel consumption and the catalog reference.\label{fig:FuelReductionPerModel}}
\end{figure}

Figure \ref{fig:FuelReductionPerModel} shows the average potential fuel reduction with each of the algorithms over every vehicle model and route type, considering the case of D1. We see how CGA2M+ generally provides fuel reductions that are more conservative than with the other two methods. The advantage is that there are no cases where the new fuel consumption is below the catalog fuel reference. Comparing "EBM\_var" with EBM, we see that the first method generally provides recommendations that decrease more the fuel consumption.

\begin{figure}[h!]
\centering
 \begin{tabular}{c@{\qquad}c@{\qquad}c}
\includegraphics[width=0.9\columnwidth]{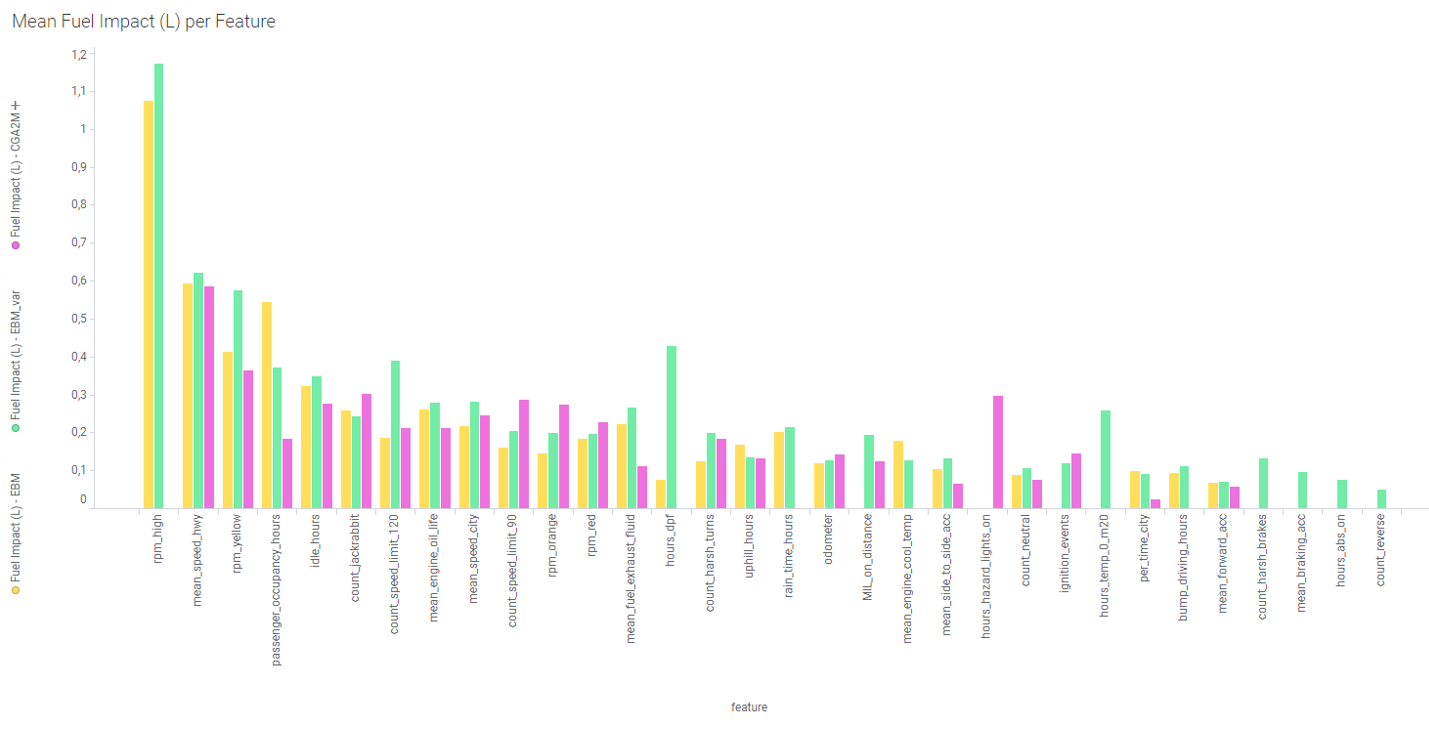}
  \end{tabular} 
  \caption{Daily mean feature fuel impact (in L) for D1, comparing the different models. Features shown appear at least within 100 vehicle-dates combinations.\label{fig:FuelImpactperFeature}}
\end{figure}

\begin{figure}[h!]
\centering
 \begin{tabular}{c@{\qquad}c@{\qquad}c}
\includegraphics[width=0.9\columnwidth]{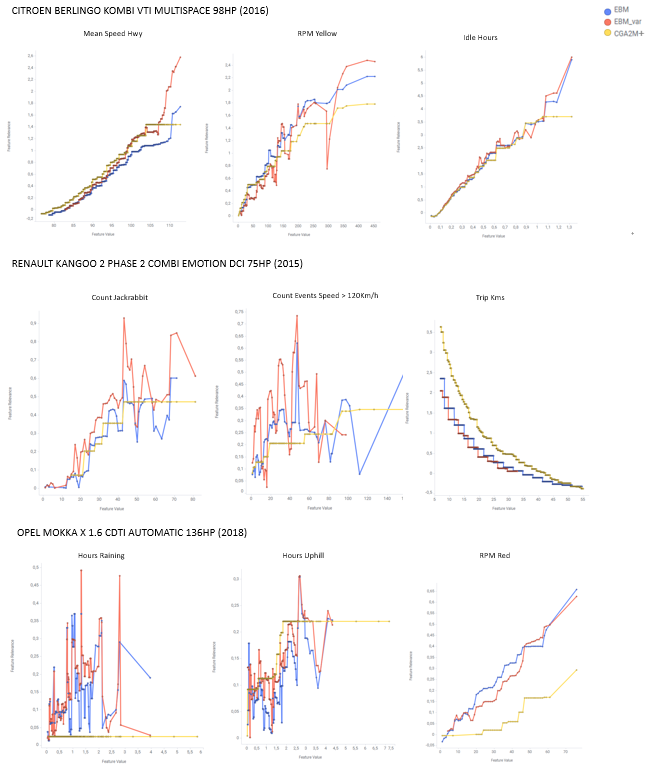}
  \end{tabular} 
  \caption{Pairplot with the relevance-values for several features considering data points for some vehicle's models only, and using the data set of D1.\label{fig:PairplotComparison}}
\end{figure}

Imposing the monotonic constraints through CGA2M+ offers more stable recommendations in exchange of covering less fuel. This can be seen in Figures \ref{fig:FuelImpactperFeature} and \ref{fig:PairplotComparison}. In Figure \ref{fig:FuelImpactperFeature} we see the daily mean feature fuel impact (in L) for D1, comparing the different models. There, we see how some of the features that have a very high impact on fuel consumption for EBM and "EBM\_var" do not appear with CGA2M+ after retrieving the explanations and applying the business rules filters. This is the case of "rpm\_high". 
Focusing on some vehicle models and some of the features with higher impact, we see in Figure \ref{fig:PairplotComparison} the relationship between feature relevance and feature values, using the data set D1. It shows how in many cases, CGA2M+ curve is below the ones from the other methods, since it needs to be monotonic. We also see how EBM and "EBM\_var" are able to extract relationships that are almost monotonic (e.g. for "Trip Kms" or "Mean Speed Hwy"), while the relationships in other cases are clearly non monotonic (e.g. "Hours Raining" or "Count Events Speed $>$ 120 Km/h).

Extending the analysis to the other data sets, and focusing on the outliers only, we get the results shown in Table \ref{table:Fuel-Saved-Explained}. We see how CGA2M+ still covers less fuel (in L) than "EBM\_var" for D1, as well as in D6 and D7, but the results differ in other cases. In D2, D3, D4, D5, D8 and D9) the fuel covered by the recommendations from CGA2M+ is equal or superior to the one from "EBM\_var". THus, even though the monotonicity constraints limits the fuel explained by CGA2M+ in some data sets, this does not happen in all of them. On average, the fuel reduced among the different data sets by CGA2M+ is 35\%.

\begin{table}[h!]
\centering
\resizebox{240pt}{!}{%
\begin{tabular}{@{}lllllll@{}}
\toprule
\textbf{Data set} &
  \textbf{Method} &
  \textbf{\begin{tabular}[c]{@{}l@{}}N data\\ points\end{tabular}} &
  \textbf{\begin{tabular}[c]{@{}l@{}}Data points\\ Explained (\%)\end{tabular}} &
  \textbf{\begin{tabular}[c]{@{}l@{}}Fuel \\ Used (L)\end{tabular}} &
  \textbf{\begin{tabular}[c]{@{}l@{}}Fuel \\ Saved (L)\end{tabular}} &
  \textbf{\begin{tabular}[c]{@{}l@{}}Fuel \\ Saved (\%)\end{tabular}} \\ \midrule
D1 & EBM      & 5772  & 80  & 28599  & 11548 & 40 \\
D1 & EBM\_var & 5772  & 80  & 28599  & 11975 & 42 \\
D1 & CGA2M+    & 5772  & 72  & 28599  & 11340 & 40 \\
D2 & EBM      & 1809  & 88  & 23139  & 12508 & 54 \\
D2 & EBM\_var & 1809  & 75  & 23139  & 9503  & 41 \\
D2 & CGA2M+    & 1809  & 86  & 23139  & 9557  & 41 \\
D3 & EBM      & 10484 & 60  & 260152 & 33211 & 13 \\
D3 & EBM\_var & 10484 & 57  & 260152 & 64060 & 25 \\
D3 & CGA2M+    & 10484 & 40  & 260152 & 84031 & 32 \\
D4 & EBM      & 1944  & 77  & 14301  & 2328  & 16 \\
D4 & EBM\_var & 1944  & 79  & 14301  & 3835  & 27 \\
D4 & CGA2M+    & 1944  & 79  & 14301  & 4846  & 34 \\
D5 & EBM      & 724   & 41  & 6816   & 1510  & 22 \\
D5 & EBM\_var & 724   & 41  & 6816   & 1567  & 23 \\
D5 & CGA2M+    & 724   & 41  & 6816   & 1625  & 24 \\
D6 & EBM      & 2003  & 96  & 11157  & 4164  & 37 \\
D6 & EBM\_var & 2003  & 90  & 11157  & 5553  & 50 \\
D6 & CGA2M+    & 2003  & 44  & 11157  & 3282  & 29 \\
D7 & EBM      & 949   & 52  & 41437  & 6681  & 16 \\
D7 & EBM\_var & 949   & 58  & 41437  & 9155  & 22 \\
D7 & CGA2M+    & 949   & 11  & 41437  & 7351  & 18 \\
D8 & EBM      & 349   & 98  & 3164   & 418   & 13 \\
D8 & EBM\_var & 349   & 98  & 3164   & 559   & 18 \\
D8 & CGA2M+    & 349   & 75  & 3164   & 1843  & 58 \\
D9 & EBM      & 10    & 80  & 471    & 41    & 9  \\
D9 & EBM\_var & 10    & 100 & 471    & 61    & 13 \\
D9 & CGA2M+    & 10    & 50  & 471    & 178   & 38 \\ \bottomrule
\end{tabular}%
}
\caption{Vehicle-dates with anomalous fuel consumption explained by the different models on the different fleets, together with the potential fuel saved (L and \%) with the recommendations.}
\label{table:Fuel-Saved-Explained}
\end{table}

\subsection{4.5. Software used}
The main libraries used for the work done in this paper are the following: 
\begin{itemize}
    \item XGBoost \cite{xgboost_repo}
    \item LightGBM \cite{lightgbm_repo}
    \item ElasticNet \cite{scikit-learn}
    \item EBM, LIME \cite{interpret_ml_repo}
    \item Tree-SHAP \cite{lundberg2020local2global}
\end{itemize}

\subsection{4.6. Limitations of our approach}
First, the domain knowledge used needs to be expressed through business rules, but this may not be suitable for all use cases. This may be improved by using a more flexible framework to gather that apriori knowledge (p.e. using ontologies). Together with that, we only work with the individual feature relevance of each variable for building the recommendations, not considering possible pairwise terms if they exist.

Also, the results may vary if other anomaly detection techniques are considered. Also, other feature selection techniques could be applied before training the model in order to filter out more the features used for training the models.

Along with that, the vehicles considered are only diesel and petrol. The results may differ if we consider hybrid vehicles.

Finally, our approach deals with explaining fuel consumption that are outliers due to several factors. This does not account for all possible features that may affect fuel usage, it uses only a subset of them. Also, we are not dealing with every possible cause of anomalous fuel usage. There are other causes, like fuel fraud, that are not considered within the scope of our proposal, mainly because they do not take place within the data sets used.

\subsection{4.7. Future Work} 
We see two main research lines where our current research can be continued. The first one is regarding the unsupervised algorithm for anomaly detection. Within our proposal, we have used a boxplot applied over the fuel consumption of the vehicles of a same group since it directly provides a limit that helps seeing the threshold value that sets apart anomalous fuel consumption and non anomalous one. It also provides a visual limit that provides an additional insight for the users since they can see the average fuel split between inliers and outliers. However, there are other unsupervised algorithms that can be used if they are able to provide that threshold limit.

The second line is regarding the XAI metric usage. The literature propose other aspects that can be measured in terms of human-friendly explanations, and is important to both include those aspects, as well as assessing with different real users that the metrics do indeed measure that aspects.

Finally, the business domain knowledge is applied after generating the explanations, but it can also be considered during the training of the models.

\section{5. Conclusion}
We have proposed a complete process for unsupervised anomaly detection in the average fuel  consumption of the vehicles of a fleet. Anomalies are explained using Explainable Artificial Intelligence (XAI) and based on the feature relevance of several variables that may impact fuel usage. The explanations take into account domain knowledge expressed through business rules, and expressed through recommendations that are adjusted depending on two different user profiles that will use them. The process is evaluated using real-world data gathered from telematic devices connected to several industry fleets.

We have also evaluated different possibilities for building a surrogate model model that infers the relationships between the input data and the predicted fuel consumption, in order to explain later how anomalous fuel consumption could be reduced. For those surrogate models, we have used Generalized Additive Models with Explainable Boosting Machine (EBM), Constrained Generalized Additive 2 Model with Consideration of Higher-Order Interactions (CGA2M+), and a proposal with a variation over the original EBM algorithm. 

In order to compare the different surrogate model alternatives, we have performed evaluations regarding model performance (how well the model predicts the target feature), and XAI metrics, that compare the explanations generated in terms representativeness, fidelity, stability, contrastiveness and consistency with apriori beliefs. 

The evaluations show that all interpretable models yield good results in terms of model performance. For the XAI metrics, particularly for the consistency with apriori beliefs, we see that the three models provide good results by themselves, and are able to provide recommendations for up to 80\% of the anomalous instances, that could potentially lead to fuel reductions of up to 88\%, on average and on the larger data sets. The XAI metrics are also used for comparing the models between them, where we saw that in general, both CGA2M+ and "EBM\_var" provide similar or better results than EBM, while respectively solving monotonicity problems and taking into account the vehicle model information to adjust the explanations.

\subsection{CRediT authorship contribution statement} 
\textbf{Alberto Barbado}: Conceptualization, Investigation, Writing - original draft, Writing - review and editing, Visualization, Formal Analysis, Methodology. 
\textbf{Óscar Corcho}: Writing - review and editing, Supervision.

\subsection{Acknowledgements} 
This research was done following the registered patent \cite{patent2020fleet} for LUCA Fleet at Telefónica. We thank Pedro Antonio Alonso Baigorri, Federico Pérez Rosado, Raquel Crespo Crisenti and Daniel García Fernández for their collaboration.

\nocite{*} 
\printbibliography

\section{Appendices}
\subsection{A.1. Acronyms}
\begin{itemize}
    \item \textbf{ML}: Machine Learning
    \item \textbf{IoT}: Internet Of Things
    \item \textbf{AI}: Artificial Intelligence
    \item \textbf{RAI}: Responsible Artificial Intelligence
    \item \textbf{XAI}: Explainable Artificial Intelligence
    \item \textbf{FAR}: Fleet Analytical Record
    \item \textbf{SOTA}: State Of The Art
    \item \textbf{VIN}: Vehicle Identification Number
    \item \textbf{GBM}: Gradient Boosting Machines
    \item \textbf{EBM}: Explainable Boosting Machines
    \item \textbf{GAM}: Generalized Additive Model
    \item \textbf{CV}: Cross-Validation
    \item \textbf{EV}: Explained Variance
    \item \textbf{ME}: Mean Absolute Percentage Error
    \item \textbf{RMSE}: Mean Absolute Percentage Error
    \item \textbf{MAE}: Mean Absolute Percentage Error
    \item \textbf{MAPE}: Mean Absolute Percentage Error
    \item \textbf{adj-R2}: Adjusted R2
\end{itemize}

\subsection{A.2. Features involved}

\begin{itemize}
    \item \textbf{vehicle id}: Vehicle's unique ID number. Index.
    \item \textbf{date}: Date (DD/MM/YYYY). Index.
    \item \textbf{vehicle model}: Vehicle's model ID (associated to its make/model/year). Categorical Feature.
    \item \textbf{make}: Vehicle's make. Categorical Feature.
    \item \textbf{model}: Vehicle's model. Categorical Feature.
    \item \textbf{year}: Vehicle's manufacturing year. Categorical Feature.
    \item \textbf{VIN}: Vehicle identification number. Categorical Feature.
    \item \textbf{route type}: Route type associated to that date (highway, city, combined). Categorical Feature.
    \item \textbf{vehicle class}: Vehicle class associated to this vehicle (depends on its average fuel consumption; e.g. Large SUVs). Vehicle Parameters.
    \item \textbf{diesel detected}: Indicates if the vehicle is diesel or not. Categorical Feature.
    \item \textbf{duration air conditioner on}: Units=Hours. Type=Positive. Zero Reference. Vehicle Parameters.
    \item \textbf{duration lights left on}: Units=Minutes. Type=Positive. Zero Reference. Vehicle Parameters.
    \item \textbf{duration ABS on}: Time driving with traction system (ABS) activated. Units=Hours. Type=Positive. Zero Reference. Vehicle Parameters
    \item \textbf{duration with change fuel filter light on}: Units=Hours. Type=Positive. Zero Reference. Vehicle Parameters.
    \item \textbf{number of cranking events below 10V}: Units=None. Type=Negative. Zero Reference. Vehicle Parameters.
    \item \textbf{duration with diesel particulate filter on}: Units=Hours. Type=Positive. Zero Reference. Vehicle Parameters.
    \item \textbf{duration PTO}: Hours using power take-off. Units=Hours. Type=Positive. Zero Reference. Vehicle Parameters.
    \item \textbf{harsh brakes events}: Units=None. Type=Positive. Zero Reference. Driving Behaviour.
    \item \textbf{harsh turns events}: Units=None. Type=Positive. Zero Reference. Driving Behaviour.
    \item \textbf{jackrabbit events}: Units=None. Type=Positive. Zero Reference. Driving Behaviour.
    \item \textbf{mean braking acc}: Mean value for braking acceleration. Units=m/s2. Type=Positive. Driving Behaviour.
    \item \textbf{mean forward acc}: Mean value for front acceleration. Units=m/s2. Type=Positive. Driving Behaviour.
    \item \textbf{mean up down acc}: Mean value for up/down acceleration. Units=m/s2. Type=Positive. Driving Behaviour.
    \item \textbf{mean side to side acc}: Mean value (absolute) for side to side acceleration. Units=m/s2. Type=Positive. Driving Behaviour.
    \item \textbf{mean speed city}: Mean value of the speed within city. Units=Km/h. Type=Positive. Driving Behaviour.
    \item \textbf{mean speed hwy}: Mean value of the speed within highways. Units=Km/h. Type=Positive. Driving Behaviour.
    \item \textbf{rpm high}: Events with engine’s speed (RPM) equal or above 1900 and below 3500. Units=None. Type=Positive. Zero Reference. Driving Behaviour.
    \item \textbf{rpm red}: Events with engine’s speed (RPM) above 3500 and vehicle speed below 40 Km/h. Units=None. Type=Positive. Zero Reference. Driving Behaviour.
    \item \textbf{rpm orange}: Events with engine’s speed (RPM) above 3500 and vehicle speed between 40 and 80 Km/h (included). Units=None. Type=Positive. Zero Reference. Driving Behaviour.
    \item \textbf{rpm yellow}: Events with engine’s speed (RPM) above 3500 and vehicle speed above 80 Km/h. Units=None. Type=Positive. Zero Reference. Driving Behaviour.
    \item \textbf{speed events over 120 Km/h}: Units=None. Type=Positive. Zero Reference. Driving Behaviour.
    \item \textbf{speed events over 90 Km/h}: Units=None. Type=Positive. Zero Reference. Driving Behaviour.
    \item \textbf{duration ecomode on}: Units=Hours. Type=Negative. Driving Behaviour.
    \item \textbf{ignition events}: Units=None. Type=Positive. Driving Behaviour.
    \item \textbf{duration speed control}: Units=Hours. Type=Negative. Driving Behaviour.
    \item \textbf{count neutral}: Total events of gear position in neutral. Units=None. Type=Positive. Zero Reference. Driving Behaviour.
    \item \textbf{count reverse}: Total events of gear position in reverse. Units=None. Type=Positive. Zero Reference. Driving Behaviour.
    \item \textbf{duration extra passenger}: Time with extra passenger. Units=Hours. Type=Positive. Zero Reference. Driving Behaviour.
    \item \textbf{height}: Mean height where the vehicle was driving. Units=Meters. Type=Negative. Environment Parameters.
    \item \textbf{duration driving uphill}: Units=Hours. Type=Positive. Zero Reference. Environment Parameters.
    \item \textbf{duration idle drive}: Units=Hours. Type=Positive. Zero Reference. Driving Behaviour.
    \item \textbf{trip kms}: Distance driven. Units=Hours. Type=Negative. Environment Parameters.
    \item \textbf{per time city}: Percentage of time spent driving within city. Units=None. Type=Positive. Environment Parameters.
    \item \textbf{duration with hazard lights on}: Units=Hours. Type=Positive. Zero Reference. Vehicle Parameters.
    \item \textbf{duration oil low light on}: Units=Hours. Type=Positive. Zero Reference. Vehicle Parameters.
    \item \textbf{duration oil change light on}: Units=Hours. Type=Positive. Zero Reference. Vehicle Parameters.
    \item \textbf{duration oil change due light on}: Units=Hours. Type=Positive. Zero Reference. Vehicle Parameters.
    \item \textbf{mean engine oil temperature}: Units=ºC. Type=Positive. Vehicle Parameters.
    \item \textbf{mean transmission oil temperature}: Units=ºC. Type=Positive. Vehicle Parameters.
    \item \textbf{variation engine oil life}: Units=None. Type=Positive. Vehicle Parameters.
    \item \textbf{mean oil pressure}: Units=Pa. Type=Positive. Vehicle Parameters.
    \item \textbf{mean engine cool temperature}:  Units=ºC. Type=Positive. Vehicle Parameters.
    \item \textbf{variation coolant level}:  Units=None. Type=Positive. Vehicle Parameters.
    \item \textbf{duration with water in fuel light on}: Units=Hours. Type=Positive. Zero Reference. Vehicle Parameters.
    \item \textbf{duration engine hot light on}: Units=Hours. Type=Positive. Zero Reference. Vehicle Parameters.
    \item \textbf{hours clean exhaust filter light on}: Units=Hours. Type=Positive. Zero Reference. Vehicle Parameters.
    \item \textbf{variation fuel exhaust fluid}: Units=None. Type=Positive. Zero Reference. Vehicle Parameters.
    \item \textbf{variation fuel filter life}:  Units=None. Type=Positive. Zero Reference. Vehicle Parameters.
    \item \textbf{distance with malfunction indicator lamp (MIL) on}: Units=Meters. Type=Positive. Zero Reference. Vehicle Parameters.
    \item \textbf{total odometer}: Maximum value of the odometer. Units=Meters. Type=Positive. Vehicle Parameters.
    \item \textbf{mean tyre pressure front-left}: Units=Pa. Type=Positive. Vehicle Parameters.
    \item \textbf{mean tyre pressure front-right}: Units=Pa. Type=Positive. Vehicle Parameters.
    \item \textbf{mean tyre pressure rear-left}: Units=Pa. Type=Positive. Vehicle Parameters.
    \item \textbf{mean tyre pressure rear-right}: Units=Pa. Type=Positive. Vehicle Parameters.
    \item \textbf{mean exterior temperature}: Units=ºC. Type=Negative. Vehicle Parameters.
    \item \textbf{duration driving with $T>0$ and $T<=20$}: Units=Hours. Type=Positive. Zero Reference. Environment Parameters.
    \item \textbf{duration driving with $T>-20$ and $T<=0$}: Units=Hours. Type=Positive. Zero Reference. Environment Parameters. 
    \item \textbf{duration driving with  $T<=-20$}: Units=Hours. Type=Positive. Zero Reference. Environment Parameters.
    \item \textbf{duration raining}: Units=Hours. Type=Positive. Zero Reference. Environment Parameters.
\end{itemize}

\newpage
\onecolumn
\subsection{A.3. General flowchart}

\begin{figure}[h!]
\centering
  \begin{tabular}{c@{\qquad}c@{\qquad}c}
  \includegraphics[width=0.7\columnwidth]{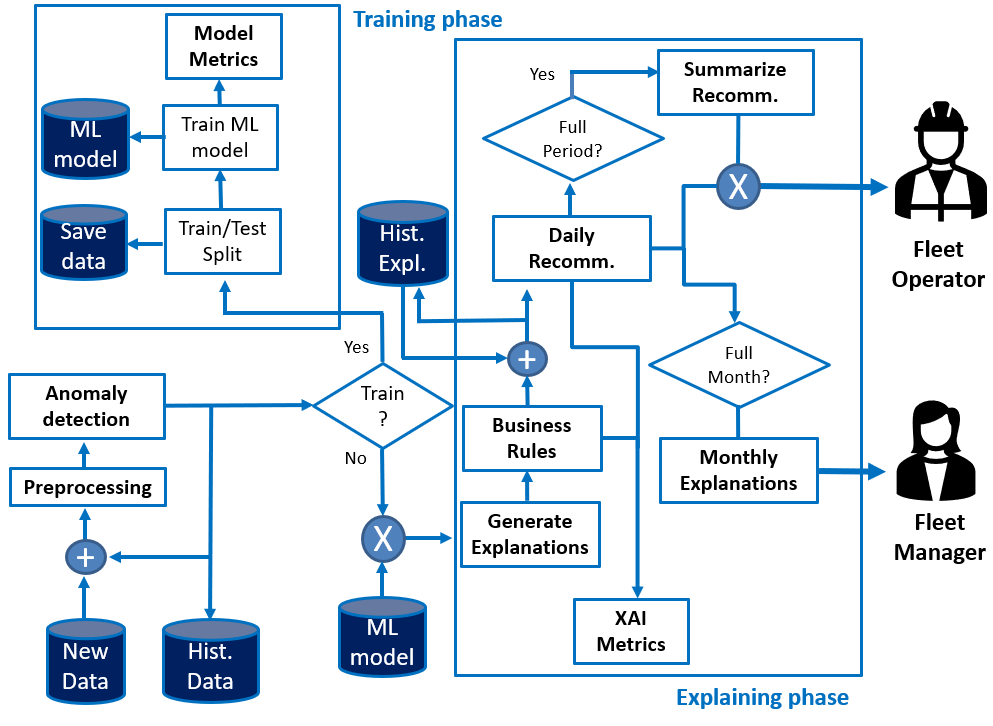}
  \end{tabular} 
  \caption{General flowchart followed by the fuel RecSys.}
  \label{fig:flowchart-fuel-recsys}
\end{figure}

\newpage
\subsection{A.4. Results}

\begin{table}[h!]
\centering
\resizebox{450pt}{!}{%
\begin{tabular}{@{}lllllllll@{}}
\toprule
\multicolumn{1}{c}{\textbf{Data set}} &
  \multicolumn{1}{c}{\textbf{Method 1}} &
  \multicolumn{1}{c}{\textbf{Method 2}} &
  \multicolumn{1}{c}{\textbf{Metric}} &
  \multicolumn{1}{c}{\textbf{Taxonomy}} &
  \multicolumn{1}{c}{\textbf{Mean 1}} &
  \multicolumn{1}{c}{\textbf{Mean 2}} &
  \multicolumn{1}{c}{\textbf{N Size}} &
  \multicolumn{1}{c}{\textbf{P-value}} \\ \midrule
Large    & EBM\_var & EBM      & n\_features\_used    & Representativeness & \textbf{6.858} & 6.429          & 5766   & 0.0     \\
Large    & CGA2M+    & EBM      & n\_features\_used    & Representativeness & \textbf{7.916} & 6.588          & 5551   & 0.0     \\
Large    & EBM\_var & CGA2M+    & n\_features\_used    & Representativeness & 6.856          & \textbf{7.742} & 5739   & 0.0     \\
Medium   & EBM\_var & EBM      & n\_features\_used    & Representativeness & \textbf{7.745} & 5.138          & 9142   & 0.0     \\
Medium   & CGA2M+    & EBM      & n\_features\_used    & Representativeness & \textbf{8.893} & 5.172          & 9028   & 0.0     \\
Medium   & EBM\_var & CGA2M+    & n\_features\_used    & Representativeness & 7.551          & \textbf{8.648} & 9768   & 0.0     \\
Small    & EBM\_var & EBM      & n\_features\_used    & Representativeness & \textbf{4.496}          & 3.635          & 685    & 0.0     \\
Small    & CGA2M+    & EBM      & n\_features\_used    & Representativeness & \textbf{8.789} & 3.884          & 431    & 0.0     \\
Small    & EBM\_var & CGA2M+    & n\_features\_used    & Representativeness & 4.682          & \textbf{8.664} & 500    & 0.0     \\
Large    & EBM\_var & EBM      & rel\_importance & Representativeness & 0.24           & \textbf{0.307} & 5800   & 0.0     \\
Large    & CGA2M+    & EBM      & rel\_importance & Representativeness & 0.2            & \textbf{0.314} & 5529   & 0.0     \\
Large  & EBM\_var & CGA2M+ & rel\_importance & Representativeness & \textbf{0.241} & 0.197          & 5797 & 0.0 \\
Medium   & EBM\_var & EBM      & rel\_importance & Representativeness & 0.202          & \textbf{0.211} & 9253   & 0.0     \\
Medium   & CGA2M+    & EBM      & rel\_importance & Representativeness & \textbf{0.255} & 0.211          & 9011   & 0.0     \\
Medium & EBM\_var & CGA2M+ & rel\_importance & Representativeness & 0.198          & \textbf{0.251} & 9900 & 0.0 \\
Small    & EBM\_var & EBM      & rel\_importance & Representativeness & 0.108          & \textbf{0.216} & 658    & 0.0     \\
Small    & CGA2M+    & EBM      & rel\_importance & Representativeness & \textbf{0.316} & 0.117          & 322    & 0.0     \\
Small    & EBM\_var & CGA2M+    & rel\_importance & Representativeness & 0.116          & \textbf{0.302} & 373    & 0.0     \\
Large    & EBM\_var & EBM      & xai\_mape                 & Precision          & 0.257          & 0.263          & 1245   & 0.316   \\
Large    & CGA2M+    & EBM      & xai\_mape                   & Precision          & 0.354          & \textbf{0.261} & 1226   & 0.003   \\
Large    & EBM\_var & CGA2M+    & xai\_mape                   & Precision          & \textbf{0.256} & 0.351          & 1251   & 0.001   \\
Medium   & EBM\_var & EBM      & xai\_mape                   & Precision          & 0.303          & 0.309          & 446    & 0.976   \\
Medium   & CGA2M+    & EBM      & xai\_mape                   & Precision          & 0.525          & \textbf{0.308} & 445    & 0.0     \\
Medium   & EBM\_var & CGA2M+    & xai\_mape                   & Precision          & \textbf{0.302} & 0.515          & 462    & 0.0     \\
Small    & EBM\_var & EBM      & xai\_mape                   & Precision          & 0.511          & 0.523          & 38     & 0.611   \\
Small    & CGA2M+    & EBM      & xai\_mape                   & Precision          & 0.626          & 0.54           & 36     & 0.735   \\
Small    & EBM\_var & CGA2M+    & xai\_mape                   & Precision          & 0.508          & 0.606          & 38     & 0.418   \\
Large    & EBM\_var & EBM      & stability\_error     & Stability          & 0.674          & 0.631          & 1719   & 0.228   \\
Large    & CGA2M+    & EBM      & stability\_error     & Stability          & 0.719          & 0.622          & 1651   & 0.131   \\
Large    & EBM\_var & CGA2M+    & stability\_error     & Stability          & \textbf{0.681} & 0.719          & 1669   & 0.024   \\
Medium   & EBM\_var & EBM      & stability\_error     & Stability          & 1.282          & \textbf{0.919} & 11526  & 0.0     \\
Medium   & CGA2M+    & EBM      & stability\_error     & Stability          & 1.645          & \textbf{0.921} & 11406  & 0.0     \\
Medium   & EBM\_var & CGA2M+    & stability\_error     & Stability          & \textbf{1.269} & 1.63           & 12570  & 0.0     \\
Small    & EBM\_var & EBM      & stability\_error     & Stability          & 5.047          & \textbf{3.669} & 748    & 0.0     \\
Small    & CGA2M+    & EBM      & stability\_error     & Stability          & \textbf{2.524} & 2.596          & 342    & 0.0     \\
Small    & EBM\_var & CGA2M+    & stability\_error     & Stability          & 3.588          & \textbf{2.574} & 371    & 0.06    \\
Large    & EBM\_var & EBM      & per\_var       & Contrastiveness    & \textbf{0.386} & 0.354          & 5307   & 0.0     \\
Large    & CGA2M+    & EBM      & per\_var       & Contrastiveness    & 0.284          & \textbf{0.379} & 5231   & 0.0     \\
Large    & EBM\_var & CGA2M+    & per\_var       & Contrastiveness    & \textbf{0.382} & 0.262          & 5217   & 0.0     \\
Medium   & EBM\_var & EBM      & per\_var       & Contrastiveness    & \textbf{0.418} & 0.252          & 8485   & 0.0     \\
Medium   & CGA2M+    & EBM      & per\_var       & Contrastiveness    & \textbf{0.417} & 0.23           & 7689   & 0.0     \\
Medium   & EBM\_var & CGA2M+    & per\_var       & Contrastiveness    & 0.37           & \textbf{0.394} & 8137   & 0.0     \\
Small    & EBM\_var & EBM      & per\_var       & Contrastiveness    & \textbf{0.268} & 0.263          & 647    & 0.008   \\
Small    & CGA2M+    & EBM      & per\_var       & Contrastiveness    & \textbf{0.54}  & 0.188          & 344    & 0.0     \\
Small    & EBM\_var & CGA2M+    & per\_var       & Contrastiveness    & 0.235          & \textbf{0.533} & 400    & 0.0     \\
Large    & EBM\_var & EBM      & per\_below           & Contrastiveness    & \textbf{0.894} & 0.874          & 31     & 0.001   \\
Large    & CGA2M+    & EBM      & per\_below           & Contrastiveness    & 0.788          & \textbf{0.874} & 31     & 0.0     \\
Large    & EBM\_var & CGA2M+    & per\_below           & Contrastiveness    & \textbf{0.894} & 0.788          & 31     & 0.0     \\
Medium   & EBM\_var & EBM      & per\_below           & Contrastiveness    & \textbf{0.792} & 0.729          & 91     & 0.0     \\
Medium   & CGA2M+    & EBM      & per\_below           & Contrastiveness    & \textbf{0.782} & 0.729          & 91     & 0.0     \\
Medium   & EBM\_var & CGA2M+    & per\_below           & Contrastiveness    & \textbf{0.792} & 0.782          & 91     & 0.0     \\
Small    & EBM\_var & EBM      & per\_below           & Contrastiveness    & 0.723          & 0.681          & 11     & 0.057   \\
Small    & CGA2M+    & EBM      & per\_below           & Contrastiveness    & \textbf{0.924} & 0.681          & 11     & 0.0     \\
Small    & EBM\_var & CGA2M+    & per\_below           & Contrastiveness    & 0.723          & \textbf{0.924} & 11     & 0.0     \\
Large        & EBM\_var       & EBM       & per\_mon       & Apriori Beliefs & 0.544        & \textbf{0.602} & 21579   & 0.0     \\
Large        & CGA2M+          & EBM       & per\_mon        & Apriori Beliefs & \textbf{1.0} & 0.605          & 22876   & 0.0     \\
Large        & EBM\_var       & CGA2M+     & per\_mon        & Apriori Beliefs & 0.543        & \textbf{1.0}   & 23415   & 0.0     \\
Medium       & EBM\_var       & EBM       & per\_mon        & Apriori Beliefs & 0.489        & \textbf{0.537} & 58344   & 0.0     \\
Medium       & CGA2M+          & EBM       & per\_mon        & Apriori Beliefs & \textbf{1.0} & 0.54           & 63232   & 0.0     \\
Medium       & EBM\_var       & CGA2M+     & per\_mon        & Apriori Beliefs & 0.49         & \textbf{1.0}   & 61237   & 0.0     \\
Small        & EBM\_var       & EBM       & per\_mon        & Apriori Beliefs & 0.549        & \textbf{0.571} & 4991    & 0.005   \\
Small        & CGA2M+          & EBM       & per\_mon        & Apriori Beliefs & \textbf{1.0} & 0.576          & 5849    & 0.0     \\
Small        & EBM\_var       & CGA2M+     & per\_mon        & Apriori Beliefs & 0.549        & \textbf{1.0}   & 5333    & 0.0   
\\ \bottomrule
\end{tabular}%
}
\caption{Hypothesis contrast for XAI metrics regarding representativeness, precision, stability,  contrastiveness, and apriori beliefs, comparing the results from EBM, EBM\_var, CGA2M+.}
\label{table:xai-metrics-contrast}
\end{table}

\pagebreak

\subsection{A5. Final Explanations and Recommendations}

\end{document}